\documentclass[journal]{IEEEtran}
\usepackage{amssymb}
\usepackage{amsmath}
\usepackage{mathtools}
\usepackage{float}
\usepackage{subfig}
\usepackage{graphicx}
\usepackage{lipsum}
\usepackage{mathtools}
\usepackage{cuted}
\usepackage{stfloats}
\usepackage{balance}

\DeclareMathOperator*{\argmin}{argmin}

\DeclareMathOperator{\Tr}{Tr}

\begin{document}
%
\title{Learning-based Regularization for Cardiac Strain Analysis with Ability for Domain Adaptation}
%
%
%

\author{Allen Lu*, Nripesh Parajuli, Maria Zontak, John Stendahl, Kevinminh Ta, Zhao Liu, \\ Nabil Boutagy, Geng-Shi Jeng, Imran Alkhalil, Lawrence H. Staib, \\ Matthew O'Donnell, Albert J. Sinusas, James S. Duncan
\thanks{This work was supported in part by the National Institute of Health (R01HL121226 and T32HL098069).}%
\thanks{*A. Lu is with the Department of Biomedical Engineering, Yale University, New Haven, CT, 06511 USA e-mail: (allen.lu@yale.edu).}%
\thanks{N. Parajuli is with the Department of Electrical Engineering, Yale University, New Haven, CT 06511 USA}%
\thanks{M. Zontak is with the College of Computer and Information Science, Northeastern University, Seattle, Washington, USA}%
\thanks{G. Jeng, M. O'Donnell are with the Department of Bioengineering, University of Washington, Seattle WA 98015 USA}%
\thanks{I. Alkhalil, N. Boutagy, Z. Liu are with the Department of Diagnostic Radiology, Yale University, New Haven, CT 06511 USA}%
\thanks{K. Ta, is with the Department of Biomedical Engineering, Yale University, New Haven, CT, 06511 USA}%
\thanks{J. Stendahl and A. J. Sinusas are with the Department of Internal Medicine and Diagnostic Radiology, Yale University, New Haven, CT 06511 USA}%
\thanks{L. H. Staib and J. S. Duncan are with Departments of Diagnostic Radiology, Biomedical Engineering, and Electrical Engineering, Yale University, New Haven, CT 06511 USA}%
}

%
%

\markboth{Draft Submission to IEEE Transactions on Medical Imaging}%
{Shell \MakeLowercase{\textit{et al.}}: Bare Demo of IEEEtran.cls for IEEE Journals}
%



\maketitle

\begin{abstract}
Reliable motion estimation and strain analysis using 3D+time echocardiography (4DE) for localization and characterization of myocardial injury is valuable for early detection and targeted interventions. However, motion estimation is difficult due to the low-SNR that stems from the inherent image properties of 4DE, and intelligent regularization is critical for producing reliable motion estimates. In this work, we incorporated the notion of domain adaptation into a supervised neural network regularization framework. We first propose an unsupervised autoencoder network with biomechanical constraints for learning a latent representation that is shown to have more physiologically plausible displacements. We extended this framework to include a supervised loss term on synthetic data and showed the effects of biomechanical constraints on the network's ability for domain adaptation. We validated both the autoencoder and semi-supervised regularization method on in vivo data with implanted sonomicrometers. Finally, we showed the ability of our semi-supervised learning regularization approach to identify infarcted regions using estimated regional strain maps with good agreement to manually traced infarct regions from postmortem excised hearts. 
\end{abstract}

\begin{IEEEkeywords}
Cardiac function, echocardiography, motion analysis, machine learning
\end{IEEEkeywords}

\IEEEpeerreviewmaketitle

\section{Introduction}
\subsection{Motivation}
\IEEEPARstart{I}{schemic} Heart Disease (IHD) remains a major problem in the United States. It is typically caused by a prolonged coronary artery narrowing, reducing blood oxygen supply to myocardial tissue. Without intervention, this leads to reversible tissue damage, \textit{myocardial ischemia} and eventually irreversible tissue damage, \textit{myocardial infarction}. Myocardial infarction leads to scarring, left ventricular (LV) remodeling, and dysfunction. Therefore, a reliable quantitative assessment of regional cardiac function for localization of myocardial injury is valuable for early detection and prompt, targeted interventions. A number of regional quantitative indicators including regional ejection fraction, wall thickness, and wall motion have been proposed utilizing imaging modalities that include Cardiac Magnetic Resonance imaging (CMR), nuclear imaging, and Computed Tomography Angiography (CTA). 4-dimensional echocardiography (4DE) has advantages of cost-effectiveness and being risk-free to the patient over other modality options. In this work, we focused on developing robust methods for estimation of regional myocardial strain for the left ventricle (LV) from 4DE. 

Regional myocardial strain estimation requires accurate and reliable motion tracking of the myocardium. Tracking methods typically follow image appearance or image-derived features over the cardiac cycle to produce a dense \textit{Lagrangian displacement field}, where all vectors reference a material point in the end-diastole (ED) frame. Although advantageous in terms of cost and acquisition time, quantitative measurement of displacements from 4DE can be challenging due to image artifacts such as inhomogeneity, bone shadows, and signal dropouts that cause poor motion estimation results. Therefore, intelligent regularization of the dense displacement field is a necessary step for producing more reliable strain analysis, which provides objective evaluation of regional heart health that leads to improved ability for diagnosis and targeted therapy.  

\subsection{Related Works}

\subsubsection{Intensity-Based Tracking}
Block Matching-based assumes a consistent speckle pattern that is propagated from frame-to-frame across the entire echo sequence. For a particular image patch or "block", a search region is defined in the subsequent image frame. Block-matching is applied to find the block in the subsequent frame that maximizes a similarity metric from the original patch of interest. Tissue motion is estimated by computing the center distance between the block of interest and the block in the subsequent frame with maximum similarity in the search region. These steps are performed for every voxel in the region of interest independently, which renders the resulting displacement field to be noisy both spatially and temporally. Therefore, an optional processing step is often performed, such as smoothing the displacement field \cite{horn1981determining, lucas1981iterative, lubinski1999speckle, chen20053, jia20103d}. The major drawback of this method is that, even with smoothing, the estimated displacements are often extremely noisy due to independent voxel-wise estimations.  
	
Elastic registration or Non-Rigid registration deforms a particular frame of interest from an image sequence to match a subsequent frame optimally, and as a result, produces a displacement field for the entire image. The resulting displacement field is represented by kernels such as B-splines, Thin-plate splines, or Radial Basis Functions (RBF), which implicitly constrained the displacement field to be spatially and/or temporally smooth. In contrast to block matching-based approaches estimating motion from each voxel independently, registration-based approaches simultaneously estimate the motion of each voxel via global optimization. However, this optimization problem is non-convex; therefore, it is usually solved in multiple steps. First, an affine registration step is taken to approximately align the two images. Then, multiple non-rigid registration steps are taken at coarse-to-fine scales to avoid local minima \cite{rueckert1999nonrigid}. Ledesma-Carbayo first used non-rigid registration registering neighboring frames and parameterized using cubic splines for cardiac motion analysis \cite{ledesma2001cardiac}.  Elen et al. used a B-Spline transformation model regularized with 3-D bending energy and volume conservation penalties on 4DE using mutual information as the similarity metric \cite{elen2008three}. However, registration-based methods tend to be computationally intensive due to solving for all voxel displacements jointly and the necessity for coarse-to-fine optimization. In addition, registration-based methods require explicit placement of grid points in the image, and misplaced grid points may bias myocardial motion. Heyde et al. proposed a LV-shaped coordinate system parameterized with B-splines \cite{heyde2013elastic}, but any anatomical-based coordinate systems require an accurate segmentation of the myocardium, which is also a difficult task. To regularize spatiotemporal cardiac motion, Ledesma-Carbayo et al. registered the entire 2D+time echo image sequence (i.e. ED frame to frame $f$ for all $f$ in the image sequence) globally by using a spatio-temporal B-spline mode \cite{ledesma2005spatio}. De Craene et al. extended aforementioned work to the application of 3D+time echo image sequence using a 4-dimensional spatio-temporal B-spline model regularizing velocities instead of displacements \cite{de2012temporal}. 
	
\subsubsection{Feature-Based Tracking}
In contrast to tracking intensity changes, another general approach is to extract relevant image features and track these features over the cardiac cycle. These features include image curvature \cite{papademetris2002estimation}, texture \cite{kaluzynski2001strain} or shapes and surfaces \cite{papademetris2002estimation, parajuli2016integrated, parajuli2017flow, shi2000point, lin2004generalized}. Examples of feature extraction include segmentation \cite{huang2014contour} for myocardial surfaces or extracting image curvatures \cite{shi2000point}. Then, point correspondences are estimated using various matching methods. Iterative Closest Point (ICP) iteratively finds correspondences and transforms points using a least squares transformation \cite{rusinkiewicz2001efficient}. Robust Point Matching (RPM) performs a similar procedure as ICP but includes fuzzy assignment and simulated annealing \cite{rangarajan1997robust}. Generalized Robust Point Matching (GRPM) extends RPM to include feature distances in addition to Euclidean distance \cite{lin2004generalized}. Feature distances such as Euclidean distance \cite{parajuli2016integrated}, surface features differences \cite{parajuli2016integrated, shi2000point, papademetris2002estimation} or image appearance dissimilarity \cite{parajuli2017flow} may be used.  Shi et al. proposed matching surface points using shape curvature as features to track and computed confidence measures based on uniqueness and correctness of the match \cite{shi2000point}. Papademetris et al. used a similar curvature-based shape tracking approach and then regularized the displacements by modeling myocardial motion with a linear elastic model \cite{papademetris2002estimation}.
The previously mentioned methods focus on spatial regularization and all lack temporal coherence. Parajuli et al.  \cite{parajuli2016integrated} imposed temporal regularization by modeling cardiac motion as the shortest path through a graph. The points of the graph are surface vertices, and the edge weights are defined based on spatial and feature distance among temporal neighborhood points. This work was extended to impose spatiotemporal constraints \cite{parajuli2017flow}. However, feature extraction (e.g. myocardial segmentation) is a challenging problem, especially for echocardiography. Therefore, performance of all surface tracking methods is limited by accuracy of segmentation method. In addition, myocardial surfaces lack shape features for capturing the torsional motion of the heart, and surface-based features, such as Euclidean distance, curvature\cite{shi2000point, papademetris2002estimation} or shape context \cite{belongie2002shape} are not rotationally invariant. Thus, torsional motion is often underestimated by surface tracking-based approaches. 

\subsubsection{Regularization Models}
Embedded in intensity and feature-based tracking methods are regularization models that enforce physiologically-plausible motion behavior, such as spatiotemporal smoothness, tissue incompressibility, and temporal periodicity. Free Form Deformation (FFD) models lay a lattice of control points on the image domain \cite{sederberg1986free, lee1996image}. These control points are displaced from their original locations, and the resulting deformation is represented by a set of polynomial basis functions such as B-splines. As a result, the local displacements enclosed within the control points are implicitly regularized. The choice of basis function determines local smoothness. For example, inclusion of higher order B-splines would allow more deformation with the raised possibility of fitting to noise. Therefore, there is an inherent trade-off between smoothness and accuracy. FFD explicitly defines a rectangular set of grid points, but myocardial deformation requires a more complex geometry. Improperly placed control points may bias displacement estimation. Extended Free Form Deformation (EFFD) models are designed to overcome issues caused by the rectangular grid from FFD \cite{coquillart1990extended}. EFFD models defines control point lattices that are adapted to the heart, such as cylindrical or anatomical \cite{ledesma2001cardiac, ledesma2005spatio, heyde2013elastic, de2012temporal, heyde2016anatomical}. These EFFD models that are especially adapted to cardiac deformation typically outperforms standard FFD models, but they are complicated to construct and requires accurate segmentation of the myocardium. 

Finite Element Method(FEM) models start by dividing myocardium using meshing techniques that facilitate incorporation of biomechanical modeling parameters. For example, Papademetris et al. \cite{papademetris2002estimation} imposed a transversely isotropic linear elastic model that incorporated a fiber model that enforced motion in myocardial directions. However, cardiac deformation in ischemic regions is not linearly elastic. Instead, the relationship between stress and segment length was determined as exponential \cite{akaishi1988non}. Furthermore, the finite element mesh was complex and difficult to construct. In contrast to FEM, Radial Basis Function (RBF)-based displacement representation do not require explicit mesh construction and are hence referred to as \textbf{mesh-free}. Choices for RBF kernels include thin-plate splines \cite{bookstein1989principal} and Gaussian kernels.  Compas et al. specifically used the computationally advantageous Compactly Supported RBF (CSRBF) for both displacement field representation and integration of shape and speckle tracking for strain analysis \cite{compas2014radial}. Parajuli et al. extended this work to incorporate a sparsity penalty for data-driven selection of RBF centers \cite{parajuli2015sparsity}. 

The aforementioned regularization models use manually crafted features that may not adhere to typical cardiac motion patterns. Furthermore, they impose spatial regularization only, and extension to spatiotemporal regularization is non-trivial. Utilizing the assumption that a well-regularized displacement field has a sparse representation, we proposed learning a sparse dictionary representation. The learned dictionary was then used for sparse coding of noisy or corrupted displacement estimates to recover the true displacements. However, sparse coding of low error trajectories resulted in quantization errors. To address this, trajectories were classified as either high or low error, and dictionary regularization was imposed on high-error trajectories only. This reduced the bias to low error trajectories when imposed with the learned dictionary but limited regularization effectiveness \cite{lu2017dictionary}.

In order to address this limitation, we extended the dictionary learning-based regularization to a supervised learning framework with a feed forward neural network (FFNN) for spatiotemporal regularization of noisy displacements. This regularization function was learned by supplying 4D Lagrangian displacement patches to a Multi-Layered Perceptron (MLP) network. We then showed the ability of this framework to generalize to various tracking methods. We also proposed combining complementary tracking methods using a multi-view learning model and showed further improved tracking and strain estimation performance. Finally, we applied the multi-view network to in vivo data and showed plausibility for domain adaptation \cite{lu2017learning}.

\subsection{Key Contributions}
	\begin{figure*}[!b]
		\begin{equation}
		\resizebox{.9\hsize}{!}{$\rho_{xyz}' = \frac{\sum_i\sum_j\sum_k W_{ijk} [I_t(x+i, y+j, z+k)I^*_{t+1}(x+l_x+i, y+l_y+j, z+l_z+k)]}{[\sum_i\sum_j\sum_k W_{ijk}|I_t(x+i, y+j, z+k)|^2]^{\frac{1}{2}} [\sum_i\sum_j\sum_kW_{ijk} |I^*_{t+1}(x+l_x+i, y+l_y+j, z+l_z+k)|^2]^{\frac{1}{2}}}$}
		\end{equation} 
		\label{equation:ncc}
	\end{figure*}
In this paper, we now fully address the problem of domain adaptation related to our previously proposed neural network-based approach in \cite{lu2017learning}. Training a supervised machine learning model requires ground-truth, which is especially difficult to obtain for dense Lagrangian displacements. Cross-domain prediction, where a model was trained in one domain and tested in another, is possible as shown in \cite{lu2017learning}, but its performance is typically poor. This is a well-documented problem in the deep learning community, and domain adaptation, defined as training one network that generalizes to multiple data domains, is needed. Previous domain adaptation methods focused on forcing the two data domains to have indistinguishable latent representations typically trained via an adversarial process \cite{ganin2014unsupervised, tzeng2017adversarial}. However, these methods are 1.) dedicated to classification problems, 2.) difficult to train due to adversarial training, and 3.) difficult to interpret. Thus, this work is a substantial expansion of \cite{lu2017learning}, where we make the following contributions:
\begin{itemize}
	\item Develops a complete approach for supervised regularization base on an autoencoder design
	\item Presents a novel semi-supervised neural network framework with biomechanical constraints for displacement regularization and domain adaptation.
	\item Validation of proposed methods on in vivo data with implanted sonomicrometers.
	\item Illustrates the promise of proposed method for identifying injury zones using estimated regional strain maps.
\end{itemize}

\section{Feed-Forward Neural Network Learning}

\subsection{Tracking Methods for Initial Displacement Estimation}
In this work, we utilize the following methods for producing initial noisy estimates of the displacement field for regularization. We chose three different representative methods: Radio-frequency-based Block Matching (RFBM), Flow Network Tracking (FNT), and Elastic Registration with Free Form Deformation model (FFD). These algorithms are described in the following sections. 

\subsubsection{RF-based Block Matching}

We utilized the RF-based block matching (\textbf{RFBM}) algorithm from Chen et al \cite{chen20053} as an input to our proposed framework. This method is performed on phase-sensitive radio-frequency (RF) images, which precede the log-compression and envelop detection steps and are complex valued. As a result, additional intensity-level information was retained for tracking in contrast to B-Mode images, which are filtered for enhanced visualization. RFBM performs tracking in the natural spherical ultrasound coordinate system that spans axial (in the direction of ultrasound beam), lateral, and elevational directions. For each voxel, a $M \times N \times K$ block around the voxel was defined, and RFBM searched for its reappearance in the next 3D image frame. The similarity between blocks was measured by a complex normalized cross-correlation function (NCC) defined in Equation 1, where $\rho_{xyz}'$ is the normalized 3D correlation coefficient at voxel $x$, $y$, $z$ as a function shifts in 3 directions $l_x$, $l_y$, $l_z$. $I_t$ and $I_{t+1}$ are the successive blocks at time $t$ and $t+1$ for comparison. Search for the block with highest NCC was confined within a local search region in the successive frame. To ensure spatial smoothness, a spatial smoothness filter was applied to the computed NCC map (i.e. NCC value for each voxel). Finally, the displacement at the voxel of interest $x$,$y$,$z$ was computed as distance between the voxel and maxima of the smoothed NCC map. The sub-voxel precision displacement in the axial direction was estimated by finding the zero-crossing of the phase of complex NCC, and a second order polynomial was fitted to the voxel-level displacement field in the elevational and lateral directions. Further details of this method can be found in \cite{jia20103d}. 

\subsubsection{Flow Network Tracking}
We also utilized Parajuli et al. called Flow Network Tracking (\textbf{FNT}) as an input to our proposed framework. First, 3D surfaces of endocardium and epicardium were extracted for each image frame using the segmentation method called Dynamical Appearance Model (DAM) developed by Huang et al. \cite{huang2014contour}. DAM discriminated class appearance differences at multiple scales by finding sparse representations of image patches for each class (i.e. blood vs. myocardium). The trained dictionaries were updated on-line through the cardiac cycle leveraging spatiotemporal coherence. Chan-Vese level set functions were fitted to the discriminated classes to produce smooth myocardial surfaces \cite{vese2002multiphase}. FNT sampled points from the extracted myocardial surfaces and assigned these points as nodes on a graph, and edges were the potential paths through the graph. FNT then solved for the optimal flow across the graph given following constraints: 
\begin{enumerate}
	\item Sum of outgoing flow is less than equal to one
	\item Sum of outgoing flow and incoming flow should be equal
\end{enumerate}
The above problem was solved with Linear Program (LP). The edge weights were precomputed as a function of Euclidean and feature distances among neighborhood points. The feature distances were learned by training a Siamese network that finds an optimal feature distance between two image patches. Feature distance was minimized when two patches were most similar and maximized when most dissimilar \cite{chopra2005learning}. Details of this algorithm can be found in \cite{parajuli2017flow}.

\subsubsection{Non-Rigid Registration}
We implemented the elastic registration-based method developed by Rueckert et al. \cite{rueckert1999nonrigid}. This method found an optimal transformation $\textbf{T}$ that mapped every voxel in image $I(x,y,z,t)$ at time $t$ to a corresponding voxel in reference image frame $I(x,y,z,t_0)$. This transformation can be split into two components: 

$$\textbf{T}(x,y,z) = \textbf{T}_{global} + \textbf{T}_{local}(x,y, z)$$ 

The global transformation model $\textbf{T}_{global}$ is an affine transformation that accounts for any potential global temporal deviations that may be caused by inconsistent placement of the ultrasound probe or breathing from the patient. Local deformation was modeled by $\textbf{T}_{local}$, which was described by a 3D Free Form Deformation (FFD) model based on B-splines. Due to the non-convex nature of the optimization, we utilized a coarse to fine optimization scheme. Finally, the overall cost function to be minimized was: 
$$ C = C_{similarity} + C_{smooth}$$
where $C_{similarity}$ is a similarity function between the transformed image and target image. There were a number of similarity function to use. We chose to use Squared Sum Distance (SSD) which is defined as: 
$$ SSD = \sqrt{\sum_{x,y,z}(I(x,y,z,t_0) - \textbf{T}(I(x,y,z,t)))} $$ 
for two arbitrary time points, $t_0$ and $t$. $C_{smooth}$ was a 3D bending energy penalty. The objective function was solved with Limited Memory-Broyden-Fletcher-Goldfarb-Shanno (L-BFGS) \cite{zhu1997algorithm}. 

This registration method produced a displacement field between two image frames, and we had two ways of utilizing this method to produce a 4-dimensional Lagrangian displacement field. In the first approach, we first registered adjacent frames to produce an Eulerian displacement field for each image frame. We then converted the Eulerian displacement field to a Lagrangian displacement field by temporally interpolating the displacements over time. We referred to this approach as Frame-to-Frame Registration using FFD model (\textbf{FFD FtoF}). In the second approach, we registered every frame in the cardiac sequence to the end-diastole frame. This approach directly produced a 4D Lagrangian displacement without the need for conversion. We referred to this approach as Frame 1-to-Frame Registration using FFD (\textbf{FFD 1toF}). Because FFD 1toF does not require temporal interpolation, this method does not incur propagation of error. In contrast, FFD FtoF would propagate errors through the cardiac sequence, which often results in ``drift". On the other hand, theoretically, registering adjacent image frames is an easier task due to relatively smaller deformation, but registering all frames of an image sequence to 1 reference frame requires solving for higher deformation and transformations, which is often problematic. Therefore, these two approaches are complementary. We utilized results with both of these registration approaches as part of our overall work. 

\subsection{Synthetic Data}
Our use of synthetic dataset was pivotal in not only validating the performance of our algorithms but providing ground-truth for development of learning-based approaches. The synthetic dataset contained eight volumetric cardiac sequences developed by Alessandrini et al. \cite{alessandrini2015pipeline} that covered a variety of different physiological conditions. The process of generating these synthetic images was as follows:

\begin{enumerate}
	\item Two real apical view clinical acquisitions were acquired from one health volunteer and one patient with dyssynchrony and candidate for cardiac resynchronization therapy (CRT). 
	\item The first frame of each acquisition was segmented for both the left ventricle (LV) and right ventricle (RV)
	\item The meshes were passed to a motion simulator i.e. the Bestel-Clement-Sorine (BCS) Electro-mechanical simulator
	\item As the parameters of the simulators are varied, different physiological cases were created, and the volume meshes obtained from the simulations were used as ground-truth
	\item Independently, sparse demons was applied to the two real recordings to produce 4-dimensional contours for the entire image sequences. These 4D contours were further edited by an experienced cardiologists. 
	\item The set of 4D landmarks were used to spatiotemporally align the real clinical image sequences with the simulated image sequences. 
	\item To simulate local echogenicity, scattering amplitude from the real recordings were sampled to create a scatter map and input into an ultrasound simulation environment (COLE) \cite{gao2009fast} to produce the final simulated ultrasound image sequences.
\end{enumerate}
These synthetic image sequences incorporated realistic ultrasounds features that simulated the difficulty in tracking ubiquitous in real ultrasound image sequences. The 8 individual sequences from the dataset simulated different physiological conditions, including one normal, 4 sequences with occlusions in the proximal (ladprox) and distal (laddist) parts of the LAD artery, left circumflex (lcx), right coronary artery (rca), and 3 sequences with dilated geometry with 1 synchronous (sync) and 2 dyssynchronous (lbbb, lbbbsmall). The non-dilated geometry image had image sizes of $224 \times 176 \times 208$ voxels with a voxel size of $0.7 \times 0.9 \times 0.6$ $mm^3$ with frame rate 23 Hz.  The dilated geometry had the same image dimensions as the non-dilated geometry but acquired with a frame rate of 21 Hz. Each image sequence contained 2250 sparse ground-truth trajectories $U_f^{sp}$ for interpolation to dense field. 

\subsection{Data Preprocessing}
\begin{figure*}[!t]
	\centering
	\subfloat[Process for extracting 4D displacement patches.]{
		\includegraphics[scale=0.3]{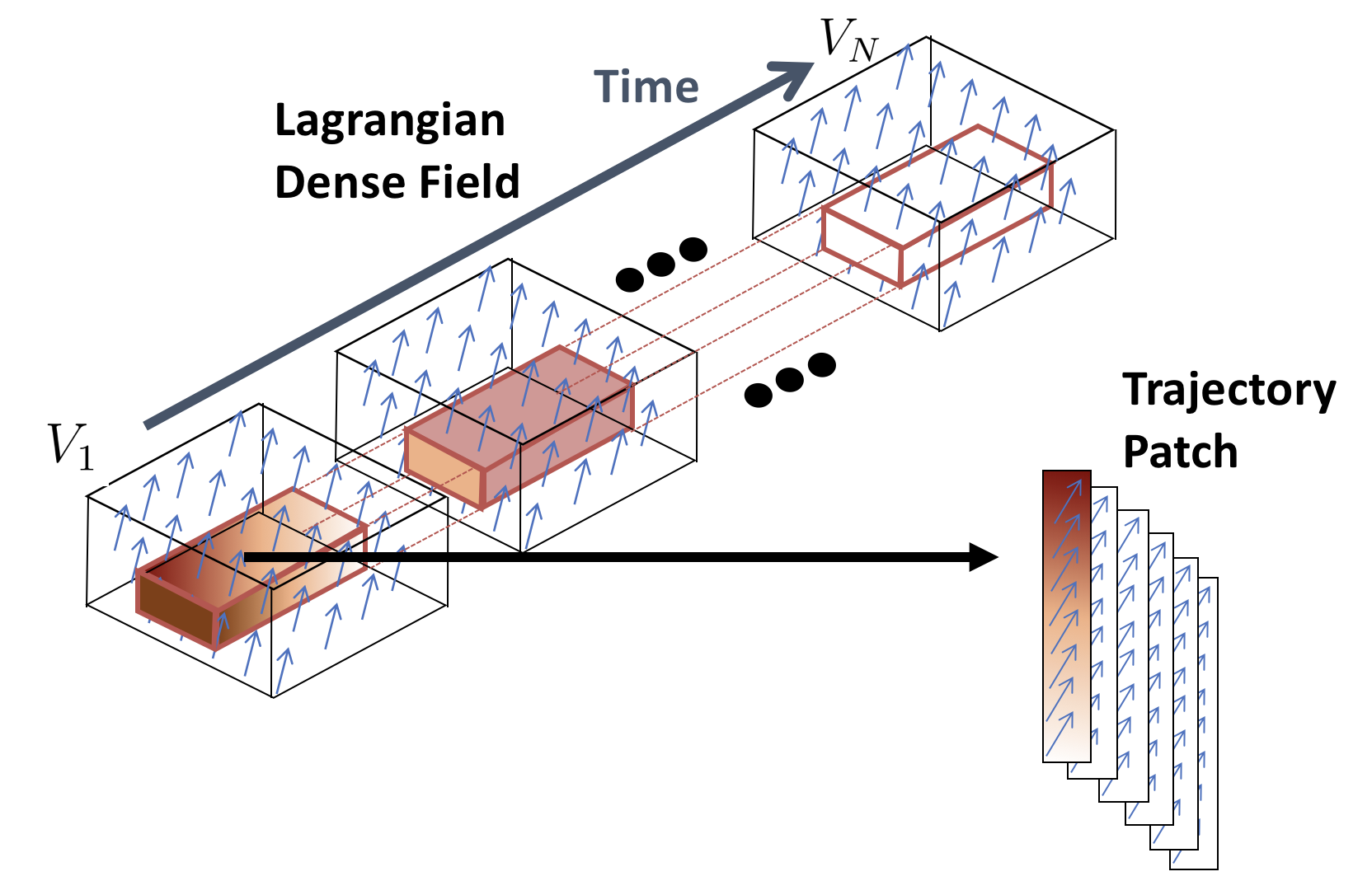}
	}%
	\hfil
	\subfloat[Process for supervised regularization learning and prediction on noisy displacement patches.]{	
		\includegraphics[scale=0.25]{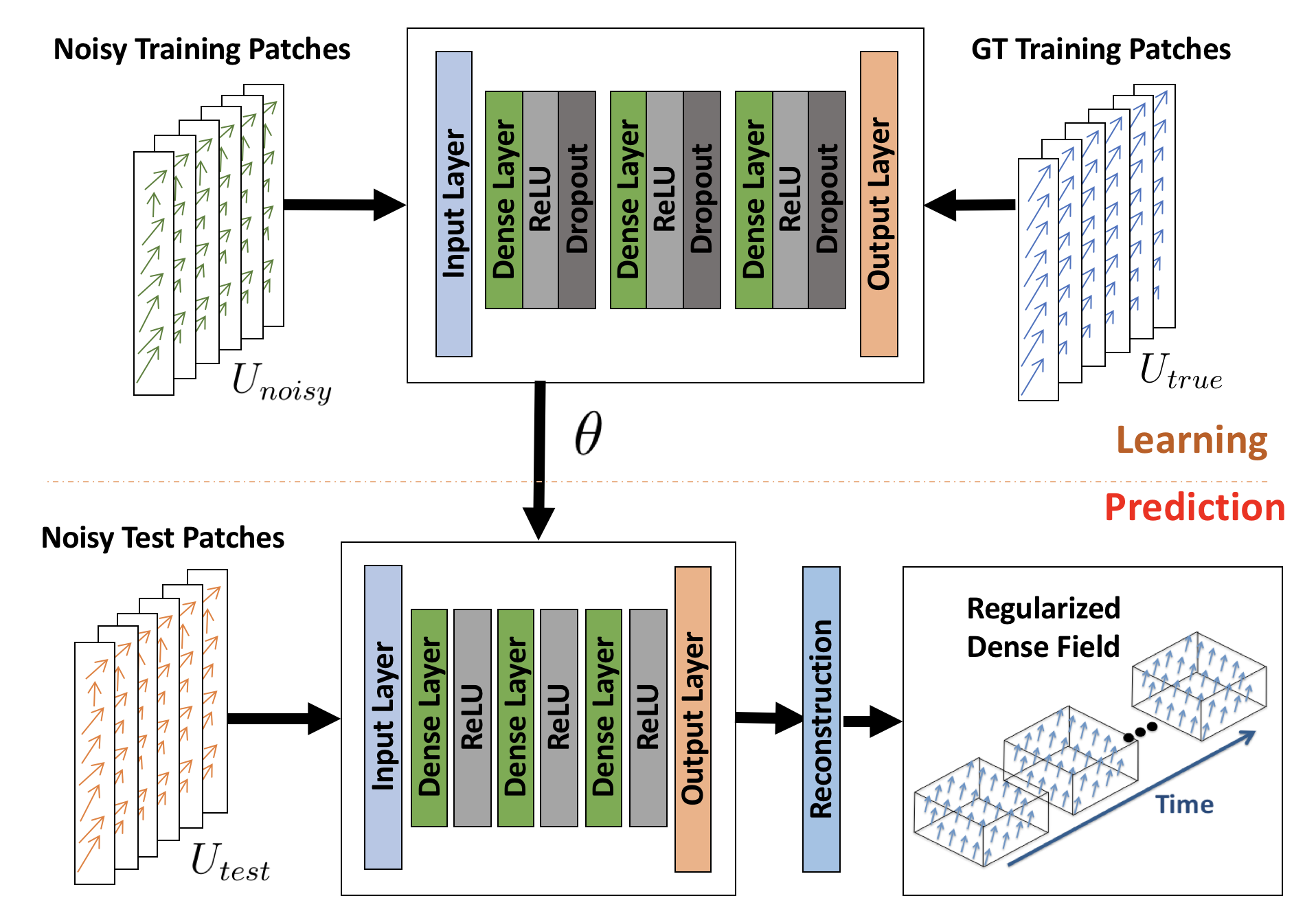}
	}
	\caption{\textbf{(a)} \emph{Extraction of 4D spatiotemporal patches from dense displacement field.} \textbf{(b)} \emph{Process diagram for training and testing of Multi-Layer Perceptron architecture.}}%
	\label{fig:supervised}%
\end{figure*}

Lagrangian dense displacement field is a 4-dimensional vector field, where the displacements at each voxel represent the motion in relation to a material point in the reference point, usually end-diastole in cardiac cycle. 4D patches were extracted from the dense field for learning spatiotemporal patterns. The sparse ground-truth trajectories in our synthetic dataset \cite{alessandrini2015pipeline} were spatially interpolated to produce dense ground-truth displacement field. Given sparse trajectories $U_f^{sp}$ for image frame $f$, we solved for frame-to-frame ground-truth dense displacement field $U_f^*$ for frame $f$ with the following objective function: 
$$ w^* = \argmin_{U_f} || Hw - U_f^{sp}||_2^2 + \lambda_1||w||_1 + \lambda_2||\nabla \cdot U_f||_2^2 $$
where $H$ defined the radial basis function kernels(RBF), $w$ are the optimal weights of RBF, and $\lambda_1$ and $\lambda_2$ are hyper-parameters for $L_1$ and divergence-free regularization terms. The resulting frame-to-frame or Eulerian displacement field $U_f$ for all $f$ were temporally interpolated with respect to material coordinates of the end-diastole frame and accumulated to produce Lagrangian displacement field for patch extraction. 

In order to learn spatiotemporal patterns, overlapping 4-dimensional patches were extracted from Lagrangian displacement fields as illustrated in Figure \ref{fig:supervised}a. The 4D patches were flattened and concatenated to form the training data. This was applied to both ground-truth Lagrangian displacement field and initial noisy displacement field estimates to form $U_{true}$ and $U_{noise}$, respectively. Corresponding pairs of $U_{noise}$ and $U_{true}$ patches were fed to the feed-forward neural network for learning the regularization function. 

\subsection{Spatiotemporal Regularization Learning}
Our goal was to learn the condition distribution that maps the noisy corrupted displacements $U_{noise}$ to $U_{true}$ by minimizing the negative log-likelihood, which is equivalent to the cross-entropy between the data distribution $P_{data}$ and model distribution $P_{model}$ \cite{goodfellow2016deep}, defined as: 
$$ C(\theta) = -\mathbb{E}_{U_{noise}, U_{true} \sim P_{data}} \log P_{model}(U_{true}|U_{noise}) $$
where $\theta$ are the parameters of the model. The specific form of $P_{model}$ determines the loss function. Assuming that $P_{model}$ has a Gaussian distribution, then the mean squared error(MSE) loss would was derived as: 
$$ C(\theta) = \frac{1}{2}\mathbb{E}_{U_{noise}, U_{true} \sim P_{data}} ||U_{true} - f(U_{noise}; \theta)||^2_2 + K$$
where $K$ is a function of the variance. While MSE loss may be used, we chose to use a Log-Cosh function, which is a smooth Huber loss function that has $L_1$ behavior for high loss, and $L_2$ behavior for small loss \cite{he2014robust}. Thus, our objective function for supervised regularization loss was: 
$$ \theta^* = \argmin_{\theta} \frac{1}{N}\sum_{i=0}^{N-1}\log\cosh\left[U_{true}^{(i)} - f(U_{noise}^{(i)}; \theta)\right] $$
where $N$ is the total number of data samples. In addition to feeding pairs of noisy $U_{noise}$ and $U_{true}$ displacement patches to the network model, pairs of $U_{true}$ and $U_{true}$ displacement patches were also fed to the model for data augmentation. In this way, the model learned to both regularize high-error displacement patches and avoid biasing low-error displacement patches via learning the identity function. The process workflow for training the MLP network and prediction were illustrated in Figure \ref{fig:supervised}b. 

\subsection{Combining Complementary Methods}
\begin{figure}[t]
	\centering
	\includegraphics[scale=0.25]{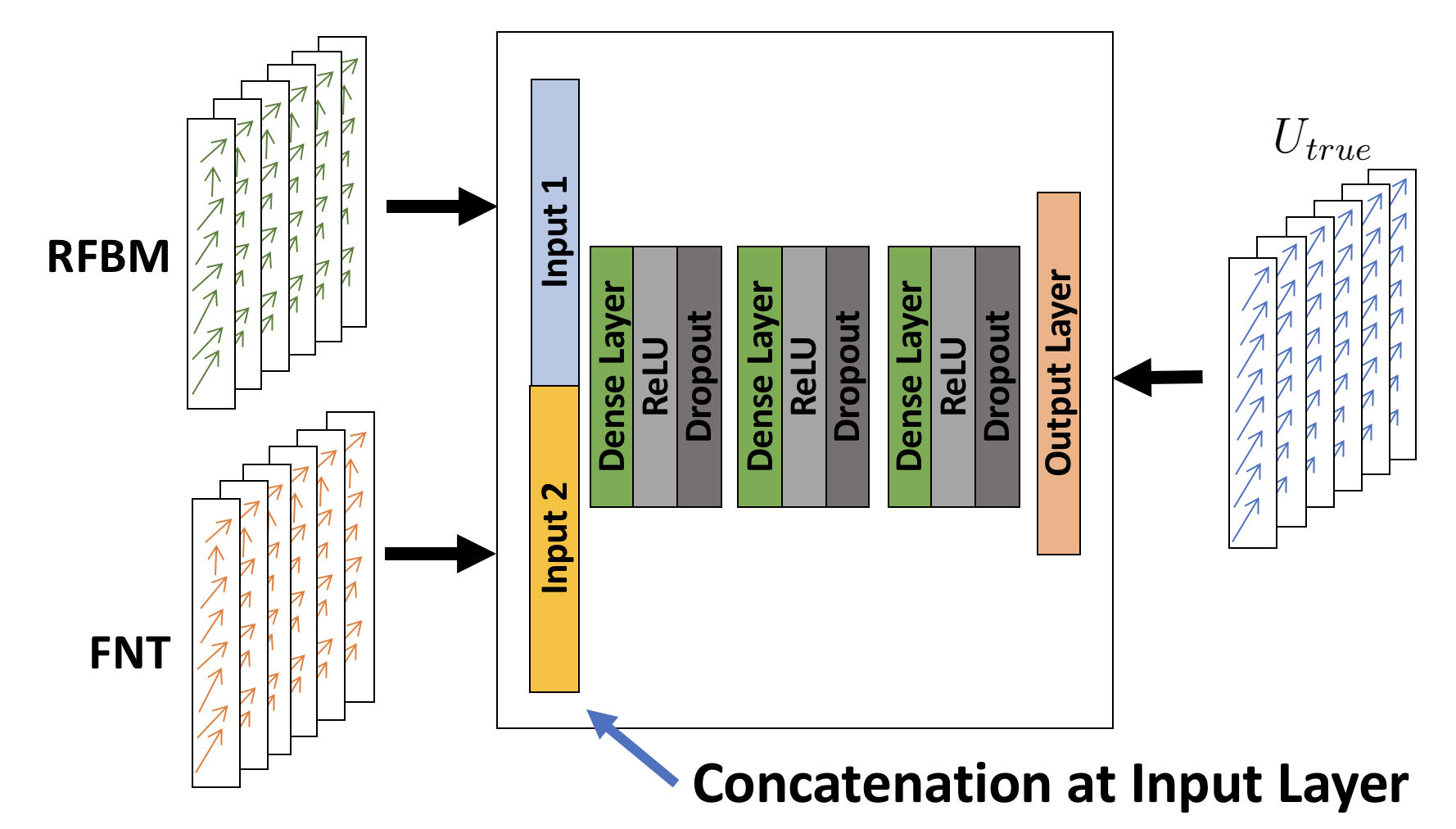}
	\caption{\emph{Multi-view learning architecture for integrating RFBM and FNT-generated displacement patches.}}
	\label{fig:multiview}%
\end{figure}
Our goal was to integrate RF-based Block Matching (RFBM) and Flow Network Tracking (FNT), two complementary tracking methods applied to inter-modal ultrasound images, for overall improved estimations. RFBM has better performance inside the myocardium but has poor performance near the boundaries due to speckle de-correlation. On the other hand, FNT tracks extracted myocardial surfaces and provides more reliable displacement estimation performance near the boundaries. Therefore, our goal was to capture the complementary nature of the two methods. 

Our approach was to learn the relationship of between RFBM and FNT. Multi-view learning \cite{sun2013survey} is a class of machine learning models that combine multiple independent sources of features and has classically been used in the medical imaging community for integrating multiple instances or views of the same object. Inspired by this, we combined the extracted displacement patches from RFBM and FNT at the input layer of our feed-forward neural network, and the network produced one set of regularized displacements from both of these sources. As a result, the network implicitly learned to weigh the inputs to produce one set of displacement estimates that captured the complementary nature of the inputs. Figure \ref{fig:multiview} illustrated the network architecture. 

\section{Domain Adaptation for Regularization of In Vivo Data}

Our previously proposed method required the availability of true displacement $U_{true}$ for learning. This ground-truth is difficult to acquire in practice. Furthermore, training a network with data from one domain (i.e. synthetic domain) and applied on another domain (i.e. in vivo domain) was challenging and usually produced poor results. 

\subsection{Autoencoder with Biomechanical Constraints}
We proposed using a biomechanically-constrained autoencoder network for learning the latent representation of noisy displacements. Autoencoders must be constrained in order to learn a useful representation, such as under-complete hidden layers, $L_1$ penalty on the parameters of the hidden layers, or sparsity constraint on the outputs of hidden layers \cite{goodfellow2016deep}. Without these constraints, the autoencoder would simply learn the identity function. In this work, we utilized prior knowledge that well-regularized displacement patches should be biomechanically plausible. Specifically, cardiac tissue deformation is near incompressible, where the volume of myocardial tissue is constant when deformed. In addition, tissue motion is approximately periodic over the cardiac cycle. Leveraging these assumptions, we introduced biomechanically-inspired constraints to the autoencoder with the following objective function:

\begin{align*} 
\theta^* = \argmin_{\theta} \sum_{i}^{N} & \{ || U_{noise}^{(i)} - U_{pred}(\theta)^{(i)} ||_2^2 + \\
\lambda_{div}\sum_{t}^T &||\Tr(\nabla U_{pred}^{(i, t)}(\theta)) ||_2^2 + \\ 
\lambda_{loop}\sum_{t}^T &||\frac{\partial U_{pred}^{(i, t)}(\theta)}{\partial t}||_2^2 \}
\end{align*}

where the first term is data adherence between $U_{noise}$ and predicted displacements $U_{pred} = f(U_{noise}; \theta)$. The second term penalized incompressibility at each frame $t$, which was measured with $L_2$ norm of divergence computed as trace of displacement gradient tensor: 
$$ \Tr(\nabla U) =\frac{\partial U_x}{\partial x} +  \frac{\partial U_y}{\partial y} +  \frac{\partial U_z}{\partial z} $$ 
where $U_x$, $U_y$, and $U_z$ were the displacement field in $x$, $y$, and $z$ directions. The third term penalized non-periodicity of cardiac motion. Summation of temporal derivatives over the temporal dimension of perfectly periodic Lagrangian displacements is zero. Thus, we penalized the $L_2$ norm of temporal derivative of Lagrangian displacements. $\lambda_{div}$ and $\lambda_{loop}$ were hyper-parameters that controlled the influence of divergence and periodicity constraints, respectively. Utilizing these constraints, the autoencoder was forced to learn a \textit{biomechanically-plausible} representation of noisy Lagrangian displacement patches.

\subsection{Semi-Supervised Learning with Biomechanical Constraints}
\begin{figure}[t]
	\centering
	\includegraphics[scale=0.25]{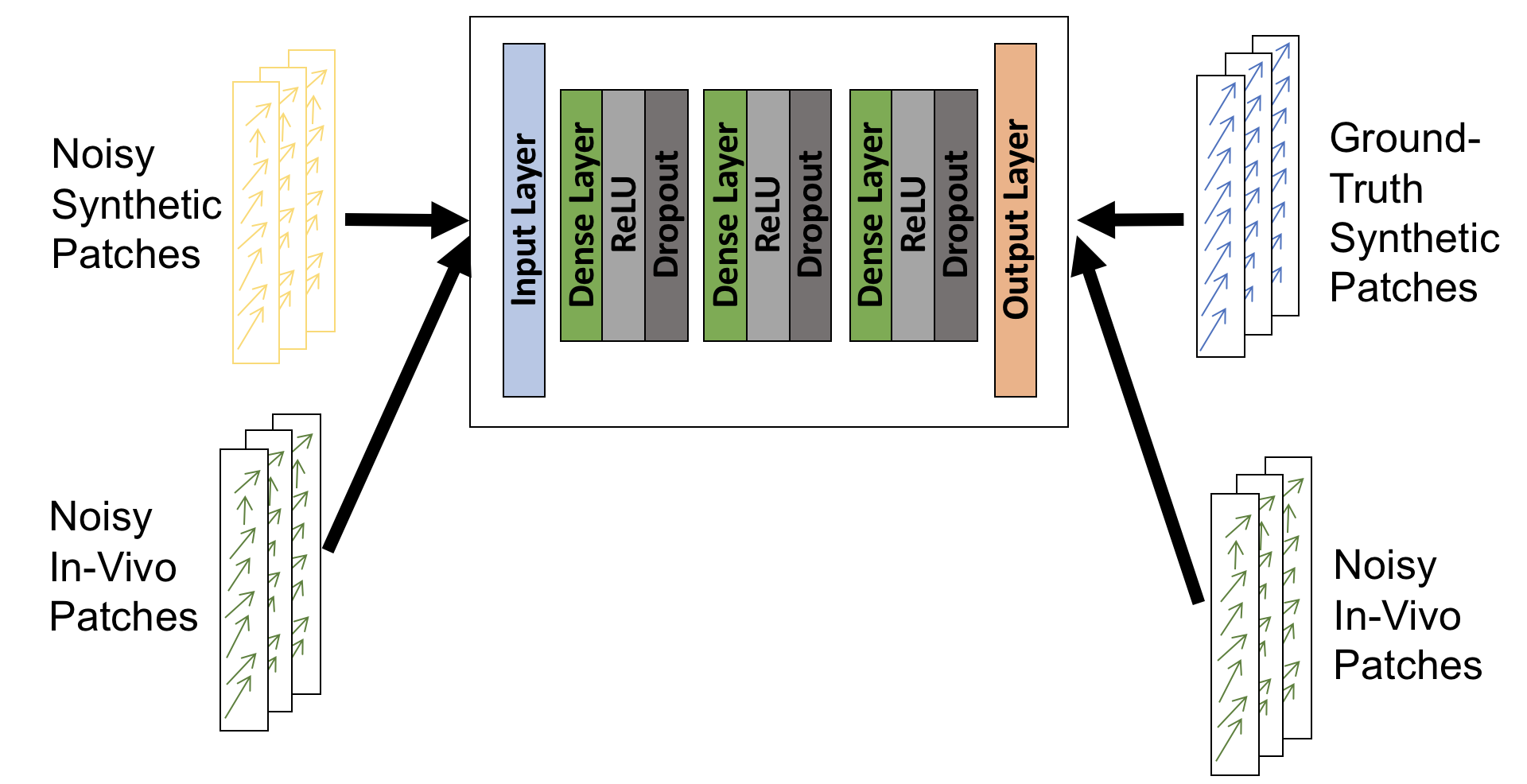}
	\caption{\emph{Semi-supervised learning architecture for integrating synthetic and in vivo displacement patches.}}
	\label{fig:semisupervised}%
\end{figure}

Finally, we proposed a method for incorporating a supervised learning term into the aforementioned biomechanically-constrained autoencoder network. We utilized the key assumption that the neural network model is able to regularize noisy displacements from both synthetic and in vivo data domains. Using this assumption, we introduced a supervised loss into the previous autoencoder objective function that minimized the $L_2$ loss between pairs of synthetic noisy displacement patches and their corresponding ground-truth displacement patches. Thus, our objective function that incorporated the supervised term(boxed) is: 

\begin{align*} 
\theta^* = \argmin_{\theta} \sum_{i}^{N} & \{|| U_{noise}^{(i)} - U_{pred}(\theta)^{(i)} ||_2^2 + \\
\lambda_{super} &\boxed{|| U_{true}^{(i)} - U_{pred}(\theta)^{(i)} ||_2^2} + \\
\lambda_{div}\sum_{t}^T &||\Tr(\nabla U_{pred}^{(i, t)}(\theta) ||_2^2 + \\ 
\lambda_{loop}\sum_{t}^T &||\frac{\partial U_{pred}^{(i, t)}(\theta)}{\partial t}||_2^2 \}
\end{align*}

This framework was illustrated in Figure \ref{fig:semisupervised}. We concatenated both supervised displacement patches (with true displacements) and unsupervised displacement patches (without true displacements) within each mini-batch for training the network. Influence of supervised term was controlled by $\lambda_{super}$. If $\lambda_{super} = 0$, then this network is simply an unsupervised autoencoder model with biomechanical constraints. We applied both unsupervised and semi-supervised neural network models on our in vivo dataset and validated with sonometric crystals. 

\section{Experimental Results}

\subsection{Synthetic Data Experiments}
\begin{table*}[!b]
	\centering
	\subfloat[Tracking Performance ]{
		\begin{tabular}{||l|c||}
			\hline
			\textbf{Methods} & \textbf{Median Tracking Error(mm)} \\ [0.4ex]
			\hline\hline
			RFBM & 1.64$\pm$1.78 \\ \hline
			RFBM-DLR & 1.48$\pm$1.55 \\ \hline
			RFBM-NNR & 0.90$\pm$0.73 \\ \hline
			FNT & 1.31$\pm$0.95 \\ \hline
			FNT-DLR & 1.28$\pm$0.86 \\ \hline
			FNT-NNR & 1.05$\pm$0.86 \\ \hline
			FFD FtoF & 1.62$\pm$1.14 \\ \hline
			FFD FtoF-DLR & 1.61$\pm$1.12 \\ \hline
			FFD FtoF-NNR & 1.16$\pm$0.80 \\ \hline
			\textbf{RBF-Comb.} & \textbf{1.46$\pm$0.91}  \\ \hline	
			\textbf{NNR-Comb.} & \textbf{0.82$\pm$0.61}  \\ \hline
	\end{tabular}}%
	\hfil
	\subfloat[Strain Estimation Performance]{	
		\begin{tabular}{||l|c|c|c||}
			\hline
			\textbf{Methods} & \textbf{Rad.(\%)} & \textbf{Cir.(\%)} & \textbf{Long.(\%)} \\ [0.4ex]
			\hline\hline
			RFBM & 21.3$\pm$72.6 & 7.0$\pm$44.0 & 5.9$\pm$45.1 \\ \hline
			RFBM-DLR & 20.2$\pm$33.9 & 4.9$\pm$19.7 & 5.7$\pm$17.5 \\ \hline 		
			RFBM-NNR & 5.9$\pm$10.7 & 2.3$\pm$2.6 & 2.4$\pm$3.4 \\ \hline
			FNT & 8.1$\pm$22.0 & 4.6$\pm$12.4 & 6.1$\pm$8.7 \\ \hline
			FNT-DL & 8.2$\pm$19.2 & 4.9$\pm$10.2 & 6.0$\pm$8.4 \\ \hline
			FNT-NNR & 4.7$\pm$11.4 & 2.6$\pm$3.4 & 2.6$\pm$3.7 \\ \hline
			FFD FtoF & 12.3$\pm$24.3 & 4.9$\pm$6.0 & 7.0$\pm$16.9 \\ \hline
			FFD FtoF-DLR& 12.1$\pm$21.7 & 4.9$\pm$5.8 & 6.9$\pm$14.9 \\ \hline
			FFD FtoF-NNR & 6.0$\pm$10.4 & 3.0$\pm$3.9 & 3.1$\pm$4.1 \\ \hline
			\textbf{RBF-Comb.} & \textbf{8.5$\pm$12.1}	& \textbf{3.7$\pm$5.3} & \textbf{3.8$\pm$5.1} \\ \hline
			\textbf{NNSTR-Comb.} & \textbf{4.0$\pm$9.8}	& \textbf{1.9$\pm$2.2} & \textbf{2.2$\pm$2.9} \\ \hline
	\end{tabular}  }
	\caption{\textbf{(a)} \emph{Median tracking error (mm) per frame compiled for all 8 studies for \textit{all} trajectories within myocardium.} \textbf{(b)} \emph{Median strain error (\%) per frame between estimated strain and ground-truth strain compiled for all 8 studies for \textit{all} trajectories within myocardium}}%
	\label{tbl:nnr_tracking_strain}%
\end{table*}

We quantitatively evaluated the performance of our algorithm on \textit{dense} trajectories (i.e. trajectory for each voxel in the myocardium). For computational efficiency, we re-sampled each voxel to 0.5 mm$^3$ with image size of $75\times 75\times 61$ voxels. We imposed a leave-one-out scheme, where we trained on 7 images and tested on the 8th image. Training patches were sampled with a stride of 2 in each direction. For normal geometry datasets (\textbf{normal, laddist, ladprox, lcx, rca}), our patch sizes were five dimensional: $5 \times 5 \times 5\times 32\times 3$ for 3 spatial directions, temporal direction, and x-y-z displacement directions. For dilated geometry datasets (\textbf{sync, lbbb, lbbbsmall}), our patch sizes were $5 \times 5 \times 5\times 39\times 3$. In total, we collected around 100,000 patches for training and 22,000 patches for testing. Our network architecture consists of 3 hidden layers with 1000 nodes per layer along with Dropout with probability of 0.2.  

Table~\ref{tbl:nnr_tracking_strain}a shows the median tracking error in mm for various different  tracking methods. We applied Dictionary Learning-based Regularization (DLR)\cite{lu2017dictionary} and Neural Network-based Regularization (NNR) to RFBM, FNT, and FFD FtoF estimates. We observed that NNR yielded significant improvements in tracking accuracy for all three methods over both initial tracked and dictionary learning-regularized trajectories (DL). 

We further compared the approach by Compas et al. \cite{compas2014radial} with our proposed multi-view network architecture for integration of surface tracking (FNT) and speckle tracking (RFBM) methods, with this method denoted as \textbf{RBF-Comb.} in Table \ref{tbl:nnr_tracking_strain}. We used RBF kernels to interpolate the sparse FNT displacements and RFBM displacements, with each sample weighted by a confidence measure. We assumed that FNT was optimal on the myocardial surfaces; thus, we assigned maximum confidence value for all FNT-derived displacements on the surfaces. For RFBM, we used NCC as a confidence measure. We compared the RBF-based combination method with our proposed learning-based integration method, where we input displacement estimates from RFBM and FNT into the multi-view learning framework denoted as \textbf{NNR-Comb.}. Furthermore,  we noticed that \textbf{RBF-Comb.}'s performance was in between the tracking accuracies of FNT and RFBM. Thus, the resulting displacement field estimate was simply an averaging between FNT and RFBM, which resulted in tracking performance that was improved from RFBM but worse than FNT. In comparison, our proposed method \textbf{NNR-Comb.} produced better tracking performance than both FNT and RFBM, suggesting that it was effective in leveraging the complementary nature of the two methods. \textbf{NNR-Comb.} produced the highest overall tracking accuracy. 

We also analyzed our performance via regional strain analysis, where we computed regional strain from dense displacement fields. The computed strain tensors were projected in clinically relevant radial (Rad.), circumferential (Cir.), and longitudinal (Long.) directions.  Strain estimation performances were shown in Table \ref{tbl:nnr_tracking_strain}b. RFBM produced incredibly high radial strain errors relative to other directions. This was because deformation was highest in the radial direction relative to other directions. Thus, RFBM needed a larger search region to capture high deformation, which meant that it was more likely for RFBM to over-fit to noise. Overall, strain performance trends mirrored that of tracking performance. NNR consistently produced improved performances over DLR-produced displacements and all three initial tracking methods. Finally, additional qualitative evaluations of strain curves and maps are shown in \cite{lu2017learning}.

\subsection{In Vivo Data Experiments}

\subsubsection{Image Acquisition Parameters}
We also acquired in vivo 4DE from anesthetized open-chest canine. These canine images were acquired using a Philips iE33 scanner (Philips Medical Systems, Andover, MA) and X7-2 probe and conducted in compliance with Institutional Animal Care and Use Committee policies. We used imaging frequency that ranges from 50-60 frames per second and typically produced around 15-30 3D volumes for each 4-dimensional sequence. 

For each study, we acquired images from the animal under 3 physiological conditions. First, we acquired a baseline image (BL) of the animal. We then introduced a total occlusion in the left anterior descending (LAD) artery for simulation of high stenosis (HO). Finally, we pharmacologically induced a stress condition by injecting the animal with relatively low-dose dobutamine ($5\mu g/ kg / min$). 

\subsubsection{Effect of Regularization Terms}
\begin{figure}[t]
	\centering
	\subfloat[$\lambda_{div} = 0, \lambda_{loop} = 0$]{
	\includegraphics[scale=0.3]{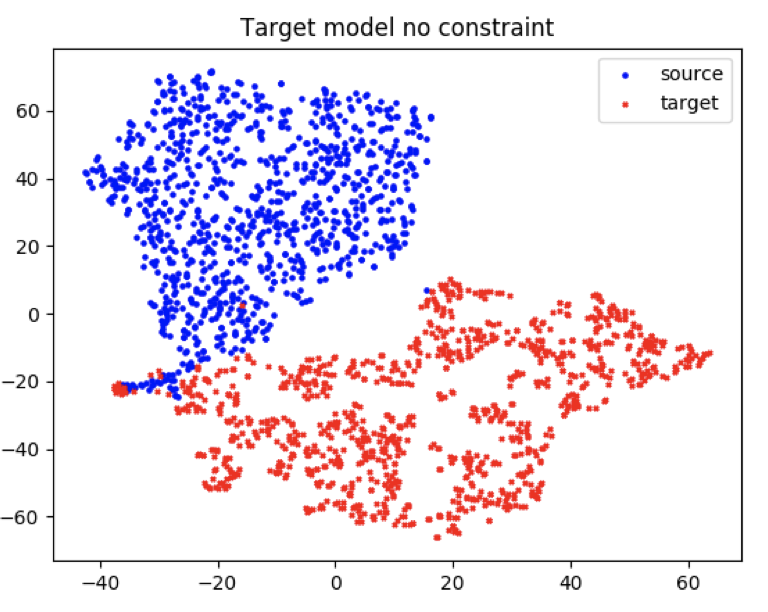}}
	\hfil
	\subfloat[$\lambda_{div} = 0.1, \lambda_{loop} = 0.1$]{\includegraphics[scale=0.3]{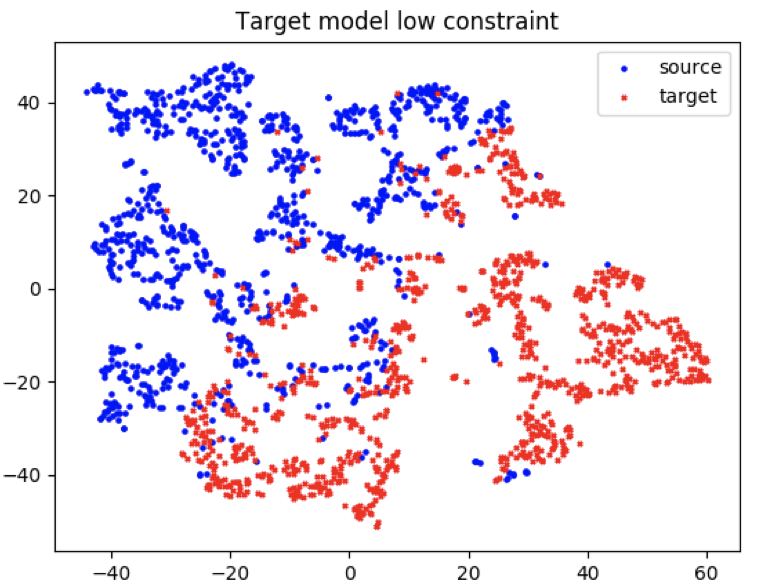}}
	\hfil
	\subfloat[$\lambda_{div} = 1, \lambda_{loop} = 1$]{\includegraphics[scale=0.3]{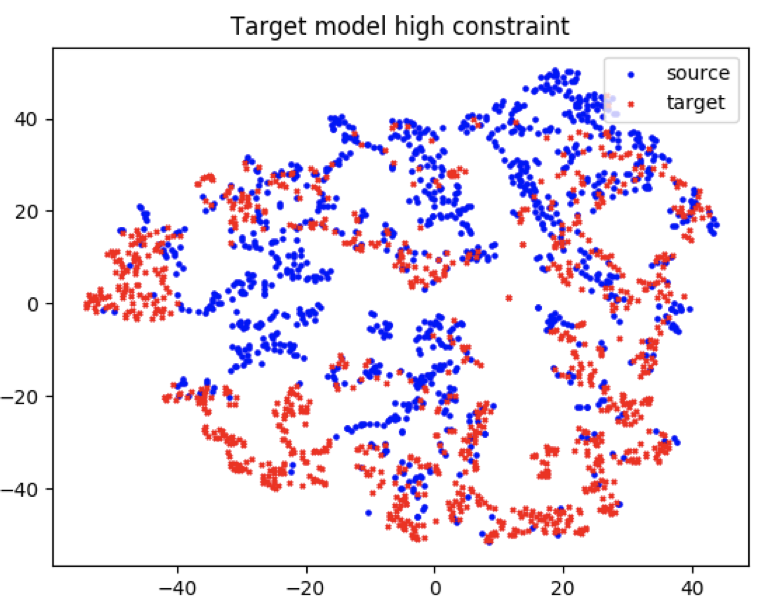}}
	\caption{(Top)\textit{t-SNE plot with no regularization} (Middle)\textit{t-SNE plot with low regularization.} (Middle)\textit{t-SNE plot with high regularization.}}
	\label{fig:tsne_constants}
\end{figure}

In this section, we analyzed the effect of the regularization parameters $\lambda_{super}$,  $\lambda_{loop}$, and  $\lambda_{div}$ on the performance of our method. Typically, machine learning models trained in one domain has worse performance when applied in another domain. Domain adaptation is the task of mitigating this issue and improving cross-domain prediction performance. We visualized the effect of these regularization parameters on domain adaptation using t-SNE, which is a nonlinear dimensionality reduction algorithm commonly used for examining the relationship of latent data representations of from different domains. In our experiment, both synthetic displacement patches and in vivo displacement patches were inputted into our semi-supervised learning model, and t-SNE was applied to the output of the \textit{last hidden layer}, which reduced the number of dimensions from 1000 to 2. We plotted the outputs from the hidden layer for three levels of regularization in Figure \ref{fig:tsne_constants}: part (a) no regularization ($\lambda_{div} = 0, \lambda_{loop} = 0$), part (b) low regularization ($\lambda_{div} = 0.1, \lambda_{loop} = 0.1$), and part (c) high regularization ($\lambda_{div} = 1, \lambda_{loop} = 1$). In part (a), the hidden layer outputs from the synthetic and in vivo data were completely separated. This indicated that the network implicitly classified the synthetic data (where outputs were true displacements) and in vivo data (where outputs were noisy displacements). As a result, the network predicted in vivo noisy displacements when the input was in vivo noisy displacements, which were not spatially smooth or periodic. This motivated the use of biomechanical regularization to force the predicted in-vivo noisy displacements to be spatiotemporally regularized, and the resulting displacements would better resemble synthetic displacements, achieving domain adaptation. Thus, we experimented with low regularization and observed a slight ``mixing" effect in part (b). We further increased regularization and observed a more significant mixing of the outputs of the two domains in part (c). This suggested that biomechanical regularization positively influenced the domain adaptation ability of the network model. 

\begin{figure*}[!t]
	\centering
	\subfloat[Sonometric Crystal Layout]{
		\includegraphics[scale=0.3]{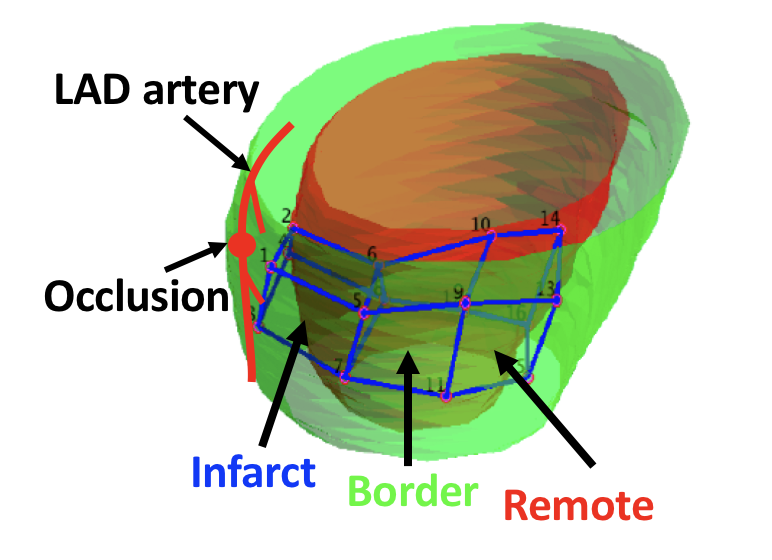}}
	\hspace{3cm}
	\subfloat[Crystal layout on transducer]{
		\includegraphics[scale=0.55]{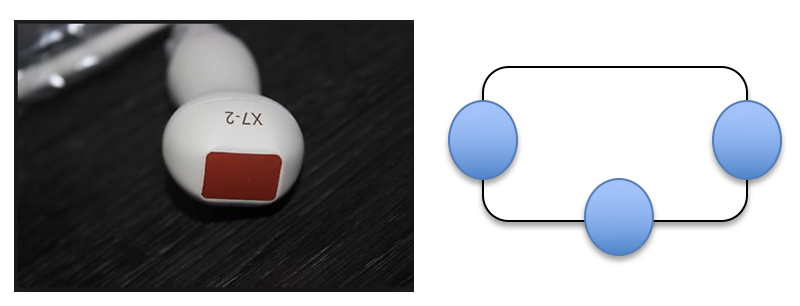}}
	\hfil
	\subfloat[Using reference crystals for mapping]{
		\includegraphics[scale=0.35]{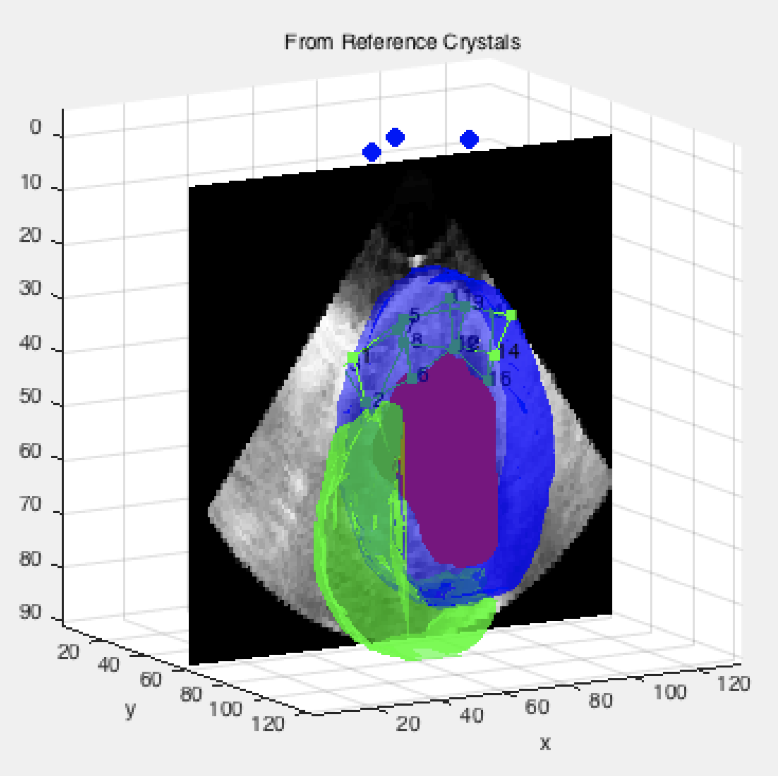}}
	\hfil 
	\subfloat[Example mapped crystals in ultrasound space]{
		\includegraphics[scale=0.35]{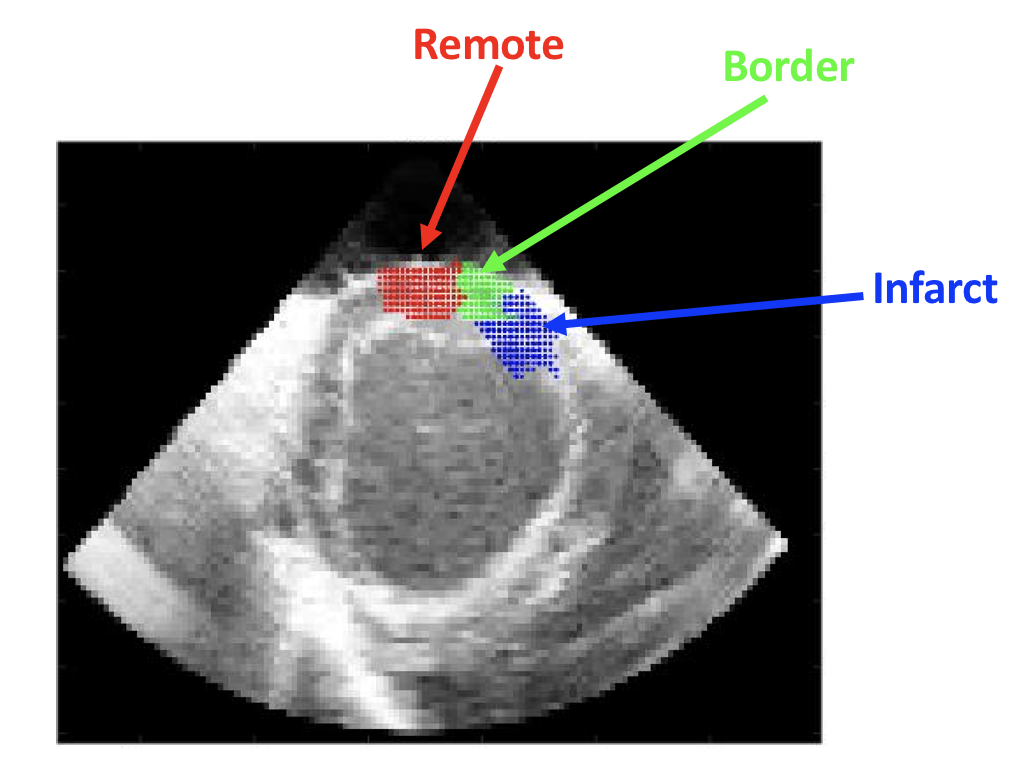}}
	\caption{\textbf{(a)} \textit{Sonometric crystals layout in relation to LAD artery} \textbf{(b)} \textit{Reference crystal on X7-2 transducer arrangement } \textbf{(c)} \textit{Mapping crystals onto ultrasound space in 3D} \textbf{(d)} \textit{Example crystals mapped on 2D image slice} }
	\label{fig:mapping3d}
\end{figure*}

\subsubsection{Sonometric Crystals}

We used sonomicrometric transducer crystals with recording instrument and software \textit{Sonosoft} and \textit{Sonoview} (Sonometrics Corporation, London, Ontario, Canada) for recording inter-crystal distances over the cardiac cycle. Cubic arrays with 3 cubes and 16 total crystals were implanted across the myocardium, where 8 crystals were placed near the endocardial surface, and 8 additional crystals were placed near the epicardial surface. One cube is approximately in the infarct zone (Infarct) near the occluded artery. One cube was away from the infarct zone (Remote). The last cube is in the middle of the two previous described cubes (Border). These cubes were shown overlaid on example LV surfaces in Figure \ref{fig:mapping3d}a. We computed strain from the 3D cubic array of crystals based on the work in \cite{waldman1985transmural}. For each cube, we defined approximately 50 tetrahedral units, or elementary units that consists of 4 out of the possible 8 vertices. Strain tensor was computed for each tetrahedral unit of the cube. Then, we computed the median strain tensors computed from all of the tetrahedral units to yield one final strain tensor. Finally, principal strain was computed via eigendecomposition of the strain tensor. 

In order to utilize the crystals-derived principal strains for validation, we used reference crystals implanted on the ultrasound transducer. We made assumptions regarding the locations of those reference crystals and solved a system of equations for the locations of 16 myocardial crystals. On the X7-2 probe, we implanted the crystals in the configuration as seen in Figure \ref{fig:mapping3d}b. The two crystals facing each other were placed approximately 28 mm apart from each other (as measured by a ruler). The third crystal was approximately 13 mm from the center of the probe surface. We assumed that these crystals were located approximately 5mm from top of the ``ultrasound fan". Based on these assumptions, we estimated locations of all 3 crystals. Then, for each myocardial crystal, the distances to the reference crystal, computed from crystal positions, should equal to the recorded distances from the crystals. We formulated this relationship as a 3 variable 3 equation problem and solved for the crystal locations.

\subsubsection{Experimental Parameters}
 For each dataset, we collected image sequences for three different physiological condition. The first condition was Baseline (BL), where the animal is healthy. The second condition was High Occlusion (HO), where we occluded the LAD artery. The third condition was High Occlusion with Dobutamine (HODOB), where we both occluded the LAD artery and stressed the animal. This allowed us to quantify the performance of our algorithm from a clinical perspective, ensuring that we captured the regional variations (i.e. strain variations across the 3 cubes) and physiological variations. Table \ref{tbl:acute_data} describes the in-vivo data and configurations for training, validation, and testing. 
 
 \begin{table}[h]
 	\centering 
 	\begin{tabular}{||l|c|c|c|c||}
 		\hline
 		\textbf{Studies} & \textbf{BL} & \textbf{HO} &\textbf{ HODOB}
& Usage
 \\ [0.4ex]
 		\hline\hline
 		DSEA08* & Available & Available & Available & Training, Test \\ \hline
 		DSEA09 & Available & N/A & N/A & Validation \\ \hline
 		DSEA10 & Available & N/A & N/A & Training \\ \hline
 		DSEA14 & Available & Available& Available & Training, Test  \\ \hline
 		DSEA15 & Available & Available & Available & Training, Test  \\ \hline
 		DSEA16 & Available &Available & Available & Training Test  \\ \hline
 	\end{tabular}
 	\caption{\emph{In-Vivo Data Overview}. DSEA08, 14, 15, 16 were used for leave-one-out testing. Training was augmented with DSEA10 and two additional sequences from DSEA08. DSEA09 was used for validation.}
 	\label{tbl:acute_data}
 \end{table}

We computed peak principal strain from each crystal cube and compared with image-derived peak principal strains.  We have 4 studies (DSEA08, 14, 15, 16) with 3 physiological conditions (BL, HO, HODOB) with 3 cubes (Infarct, Border, Remote) for each image sequence. In our leave-one-out testing scheme, we have N=36 samples for comparison. Our testing metric was Pearson correlation computed for the 36 samples. We tested our semi-supervised learning framework on Flow Network Tracking (FNT), RF-Block Matching (RFBM), Frame-to-frame non-rigid registration (FFD FtoF) and registering frame 1 to each frame (FFD 1toF). For each experiment, we extracted approximately 100,000 displacement patches from the synthetic datasets and 100,000 displacement patches from the in-vivo datasets, totaling approximately 200,000 patches. To accommodate for the increase in data, we increased the number of hidden layers from 3 (in our synthetic data experiments) to 7 for these in-vivo experiments. For each dataset, we computed correlations from peak strains estimated using the initial tracking method (\textbf{FFD FtoF, FFD 1toF, RFBM, FNT}). We also showed the computed correlations from peak strains regularized with a neural network model trained \textit{only on the synthetic dataset} (\textbf{Synthetic}). Further, we showed computed correlations with an Autoencoder with bio-mechanical regularization, which was equivalent to setting $\lambda_{super}=0$ (\textbf{Autoencoder}). Finally, we showed peak strain correlations with our semi-supervised framework (Supervised term and Autoencoder) with bio-mechanical regularization (\textbf{Semi-supervised}), where we set $\lambda_{super}=1$. 

\subsubsection{Comparison of Methods}
\begin{figure*}[!t]
	\centering
	\subfloat[Comparison of RFBM-produced and regularized peak strains for DSEA16]{\includegraphics[scale=0.3]{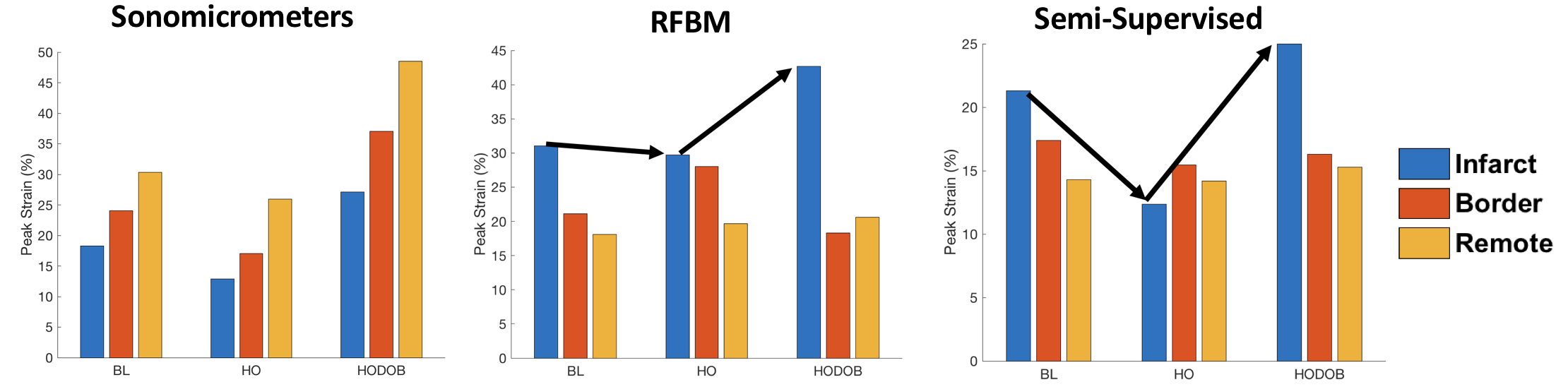}}
	\qquad \qquad \qquad 
	\subfloat[Comparison of FFD 1toF-produced and regularized peak strains for DSEA16]{\includegraphics[scale=0.3]{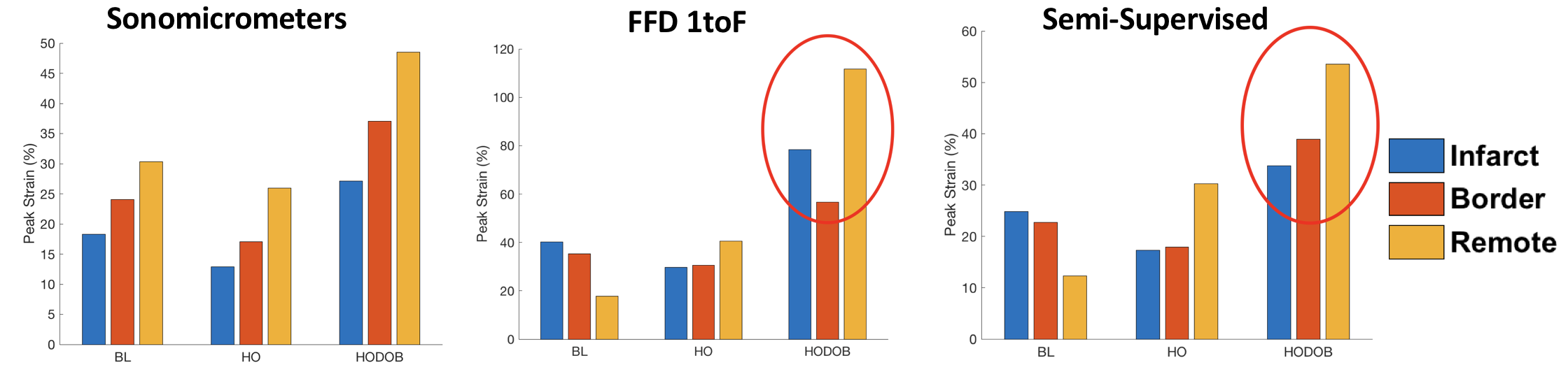}}
	\caption{\textbf{(a)} \textit{Comparison of RFBM-produced and regularized peak strains.} (b) \textit{Comparison of FFD 1toF-produced and regularized peak strains.} Both (a) and (b) show evidence of improvements in strain trends after regularization.}
	\label{fig:rfbm_ffd_example}
\end{figure*}

\begin{table}[h]
\centering 
		\begin{tabular}{||l|c|c|c|c||}
			\hline
			\textbf{Studies} & \textbf{RFBM} & \textbf{FNT} & \textbf{FFD FtoF}
& \textbf{FFD 1toF}
 \\ [0.4ex]
			\hline\hline
			Without Regularization & 0.01 & 0.17 & 0.33 & 0.60 \\ \hline
			Synthetic Model & 0.15 & 0.04 & 0.37 & 0.49 \\ \hline
			Autoencoder & \textbf{0.27} & \textbf{0.26} & \textbf{0.52} & 0.60 \\ \hline
			Semi-Supervised & 0.26 & 0.25 & \textbf{0.52} & \textbf{0.63}  \\ \hline
	\end{tabular}
\caption{\emph{Pearson Correlation between crystal and image-derived peak strains (N=36)}}
\label{tbl:acute}
\end{table}

Pearson correlations for the various tracking methods were presented in Table \ref{tbl:acute}. RFBM-produced displacements were regularized using $\lambda_{div} = 0.5$ and $\lambda_{loop} = 0.5$. As expected, RFBM without any regularization produced poor results with correlation of 0.01. The severely low correlation stemmed from the fact that principal strain captured the highest deformation and was similar to radial strain, which increased the possibility of over-fitting to noise. A model trained on synthetic data only improved the global correlation to 0.15. However, with using our unsupervised and semi-supervised learning-based regularization, we were able to capture higher correlations of 0.27 and 0.26, respectively. 

We applied our algorithm to noisy displacement estimates from FNT using $\lambda_{div} = 0$ and $\lambda_{loop} = 1$. We observed an overall increase in global correlations from 0.17 to 0.25. The relatively low correlation for FNT was due to poor segmentation results used for surface tracking. Performance of FNT or any feature tracking-based methods was heavily dependent on the accuracy of the feature extraction process. In the case, FNT relied on segmentation accuracy. In our experiments, we used segmentation method described in \cite{huang2014contour}. In this method, the end-diastole (ED) frame was manually segmented, and the resulting contours were propagated bi-directionally towards the end-systole (ES) frame. As a result, segmentation errors propagated temporally and were highest at ES frame. This was problematic for computing peak strain, which was typically from computed from ED to ES. 

We applied our algorithm to noisy displacement estimates from both non-rigid registration approaches. Both of these methods were implemented using $\lambda_{div} = 0.5$ and $\lambda_{loop} = 0.5$. For \textbf{FFD FtoF}, correlation improved from 0.33 to 0.52 for both unsupervised and semi-supervised learning approaches. We observed that FFD 1toF produced the highest correlation of 0.6 compared to FFD FtoF, RFBM, and FNT and improved to 0.63 with semi-supervised regularization. FFD 1toF produced higher correlation than FFD FtoF likely due to error of propagation significantly affecting performance of FFD FtoF. 

In summary, we observed that supervised loss term did not significantly affect peak strain correlations. This was likely due to synthetic dataset being significantly different from in-vivo dataset. First, synthetic datasets were significantly less noisy compared to the in-vivo dataset. Second, the synthetic datasets were acquired from humans, but our in-vivo datasets were acquired from canines. Third, our synthetic datasets were always oriented vertically in the image domain, but the in-vivo datasets were acquired in a variety of probe angles, which resulted in the LV being oriented at different angles. Data normalization between the two domains with augmentation to the synthetic domain should significantly improve the performance of our semi-supervised regularization approach. 

Nonetheless, we observed evidence of improvements in peak strain trends for DSEA16 in Figure \ref{fig:rfbm_ffd_example}a. We saw that peak strain in the infarct cubes were in a more clear ``V" shape in our proposed approach. Furthermore, in the high occlusion condition dataset (HO), we expected to see a positive gradient, where the ischemic cube had the lowest strain, and the remote cube had the highest strain. RFBM-derived peak strains had the opposite trend, which was implausible. After regularization, we observed closer to the ``upward" trend or positive gradient that we expected and observed from the sonomicrometer data. Figure \ref{fig:rfbm_ffd_example}b illustrated an example of correcting peak strain trend from FFD 1toF to a positive gradient in the HODOB condition after regularization. 

\subsubsection{Combining Multiple Methods}
\begin{figure*}[!t]
	\centering
	\subfloat[Comparison of FFD FtoF-produced and regularized peak strains combining FNT and FFD FtoF]{\includegraphics[scale=0.3]{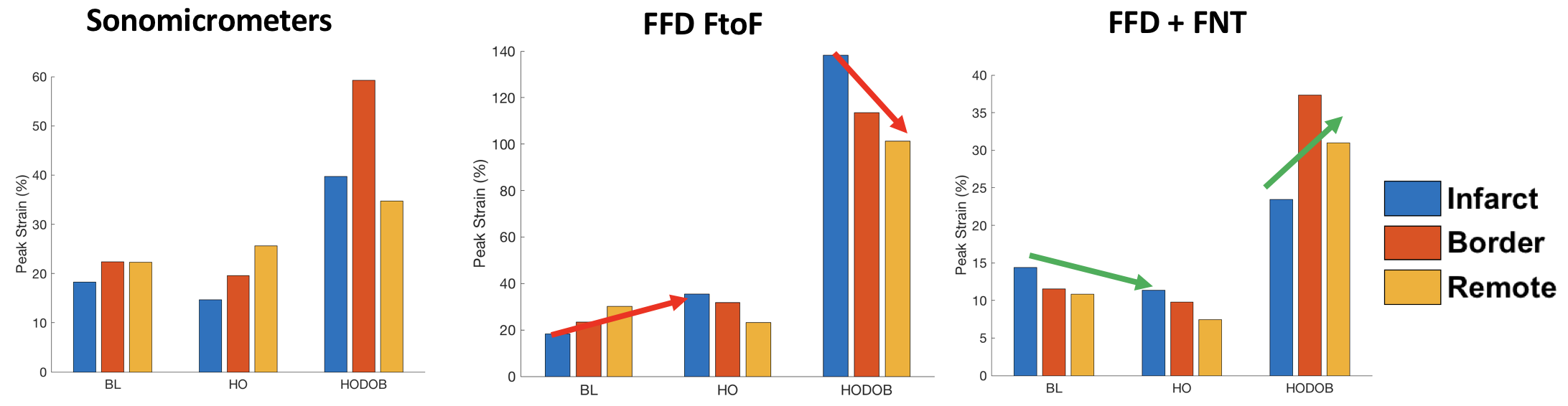}}
	\qquad \qquad \qquad 
	\subfloat[Comparison of FNT-produced and regularized peak strains combining FNT and FFD FtoF]{\includegraphics[scale=0.3]{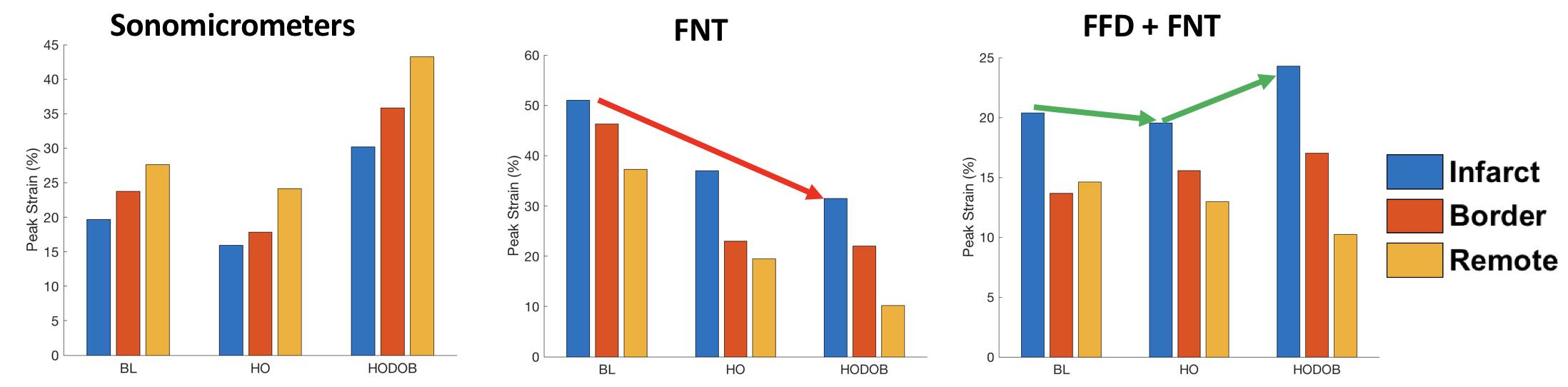}}
	\caption{\textbf{(a)} \textit{Comparison of FFD FtoF-produced and regularized peak strains.} (b) \textit{Comparison of FNT-produced and regularized peak strains.} Both (a) and (b) show evidence of improvements in strain trends after combination of both methods with regularization.}
	\label{fig:combination_example}
\end{figure*}

\begin{table}[t]
	\centering
		\begin{tabular}{||l|c||}
			\hline
			\textbf{Studies} & \textbf{Correlation} \\ [0.4ex]
			\hline\hline
			FNT + FFD FtoF Autoencoder
& 0.57 \\ \hline
			FNT + FFD FtoF Semi-Supervised
& \textbf{0.60} \\ \hline
			FFD 1toF + FFD FtoF Autoencoder
 & 0.65 \\ \hline
			FFD 1toF + FFD FtoF Semi-Supervised
& \textbf{0.67}
 \\ \hline
		\end{tabular}  
\caption{\emph{Pearson Correlation between crystal and image-derived peak strains (N=36)}}
\label{table:combined}
\end{table}

We tested integration of multiple tracking methods, where we inputted two sets of noisy displacements concatenated at the input layer and produce one set of regularized displacement output. We experimented two combinations that were thought to be promising. We first combined FNT and FFD FtoF. FNT relied on tracking surface points, but non-rigid registration tracked intensity information. Therefore, these two methods provided independent features and produced overall better performance than the individual methods. Specifically, FNT produced 0.17 correlation, and FFD FtoF produced 0.33 correlation. This combined method produced 0.57 and 0.60 for the autoencoder and semi-supervised models, respectively.

We also experimented with utilizing both FFD 1toF and FFD FtoF. FFD 1toF registered each frame in the image sequence to 1 reference frame, and in contrast, FFD FtoF registered adjacent frames in the image sequence and converted the resulting Eulerian displacements to Lagrangian displacements. Theoretically, registering adjacent image frames is easier due to smaller deformation between adjacent frames. In contrast, registering between a reference image frame (e.g. ED) to another frame in the image cycle (e.g. ES) would be harder due to high deformation between the two image frames. On the other hand, the process of converting Eulerian to Lagrangian displacements incurs a propagation of error, but FFD 1toF directly produced Lagrangian displacements that did not require this conversion. Therefore, these two methods were complementary in the temporal domain. Our network combining these two methods produced a correlation of 0.67, which was higher than correlations of 0.33 and 0.60 from the individual methods. 

Finally, we showed an example of the FNT and FFD FtoF combination in Figure \ref{fig:combination_example}. The first column showed the crystal-derived strains. The middle column showed peak strains from FFD FtoF (top) and FNT (bottom) for DSEA15 and DSEA08, respectively. FFD produced higher peak strain for HO condition in the infarct zone than that in BL, which contradicted crystal-derived peak strains. Also, FNT produced peak strains in the infarct zone that completely contradicted the crystal-derived peak strains and was due to poor segmentation results. Last column showed the peak strains estimated by the combined method. We noticed that peak strain trends had better agreement to corresponding crystal-derived strains in both DSEA15 and DSEA08. Overall, our combined approach had a boosting effect, where two sets of independent displacement estimates were jointly utilized to produce better overall peak strain estimations.

\subsubsection{Comparison to Infarct Zones}

\begin{figure*}[!t]
	\centering
	\subfloat[Apex]{\includegraphics[scale=0.2]{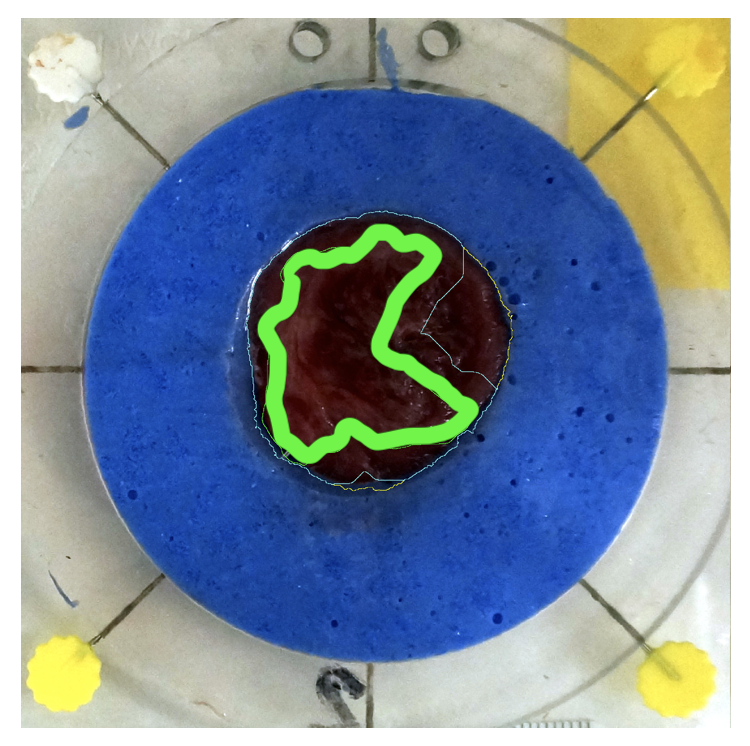}}
	\hfil
	\subfloat[Base]{\includegraphics[scale=0.2]{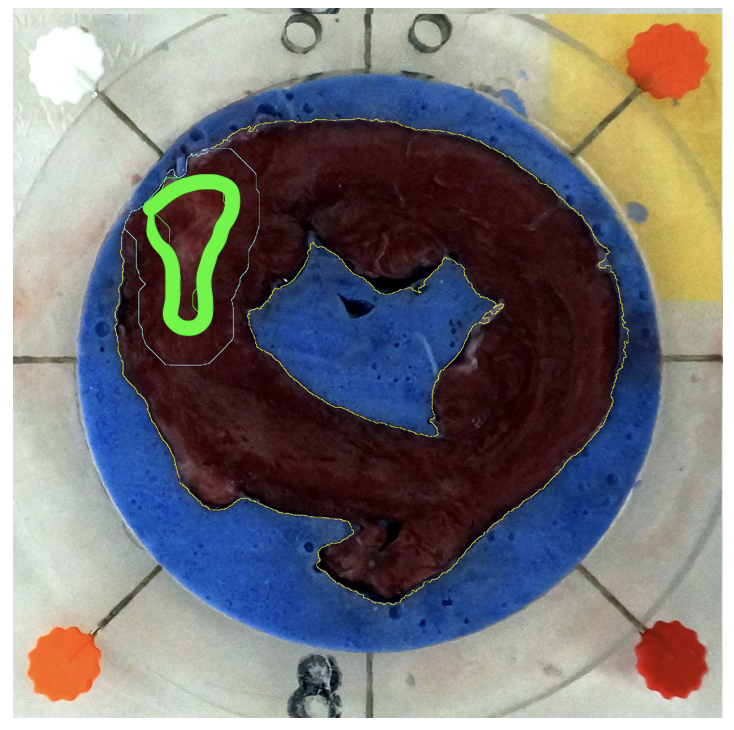}}
	\hfil
	\subfloat[3D Tracings]{\includegraphics[scale=0.2]{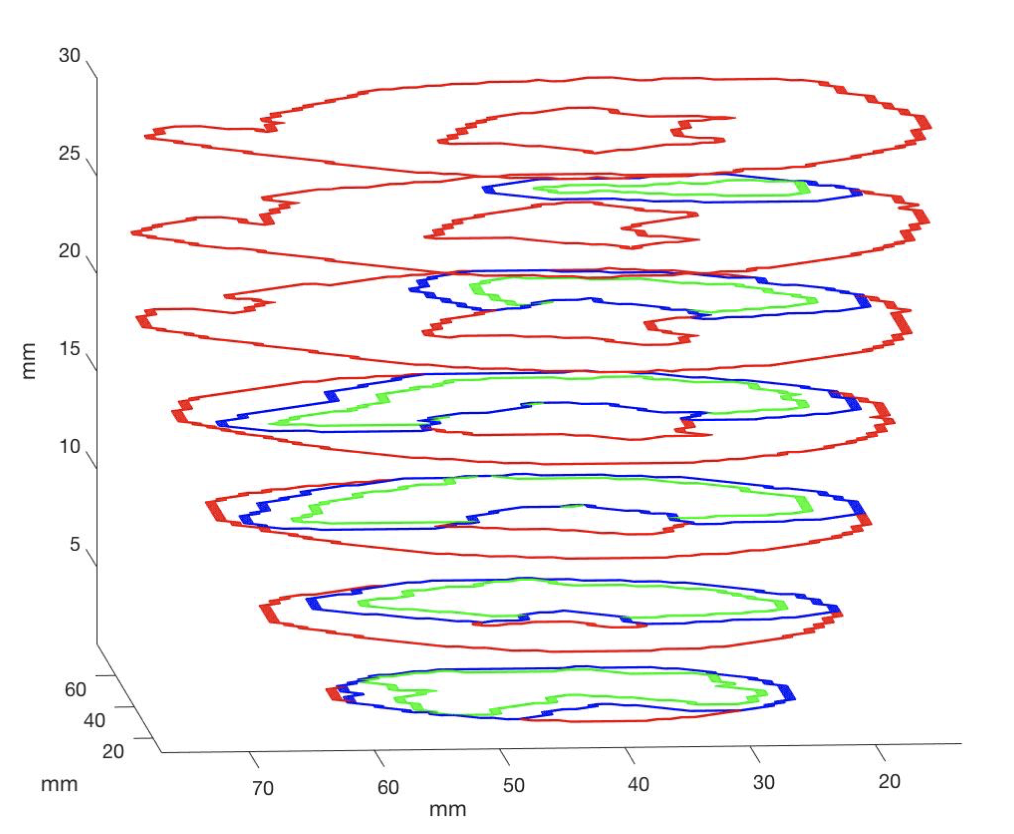}}
	\hfil
	\subfloat[LAD Infarct]{\includegraphics[scale=0.25]{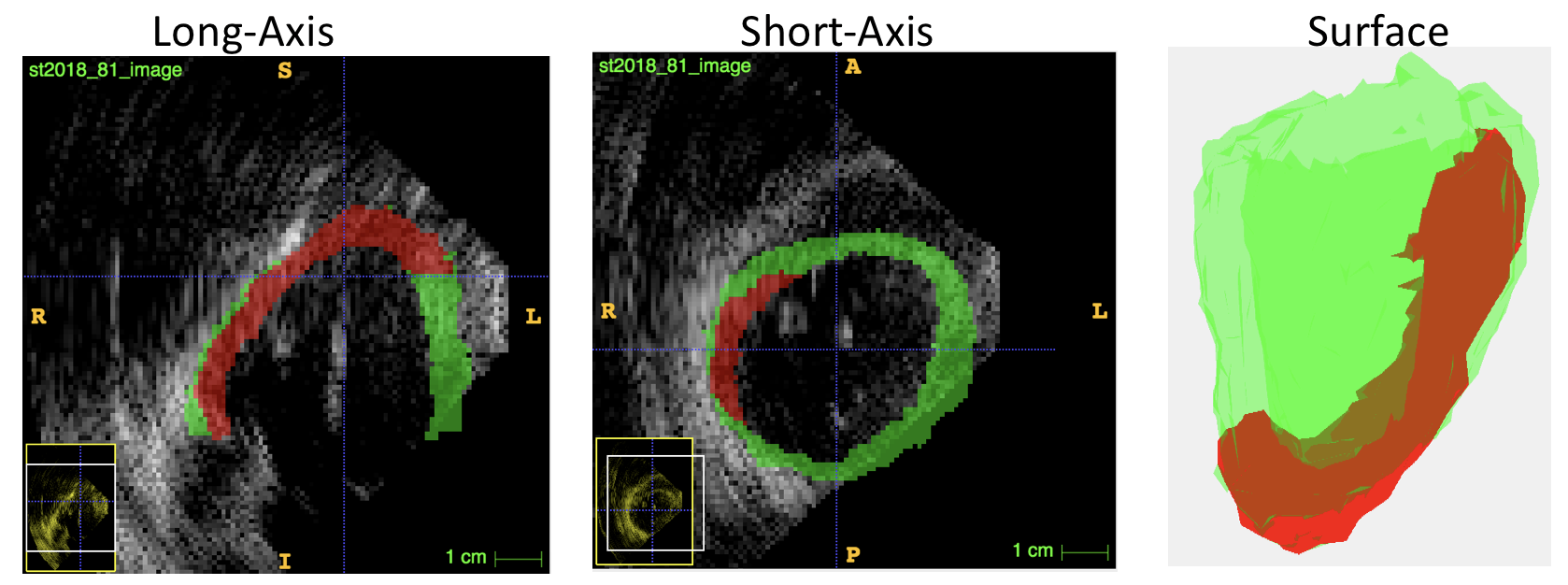}}
	\caption{\textit{LAD infarct manual tracings from a postmortem excised LV.} Part (a) shows a cross-section of the LV near the apex. Part (b) shows a cross-section of the LV near the base. Part (c) shows contours in 3D of traced LV with infarct (green) and peri-infarct (blue) zones. Part (d) shows manually traced LAD infarct onto ultrasound space.}
	\label{fig:lad_trace}
\end{figure*}

\begin{figure*}[!t]
	\centering
	\subfloat[DSEC05]{\includegraphics[scale=0.3]{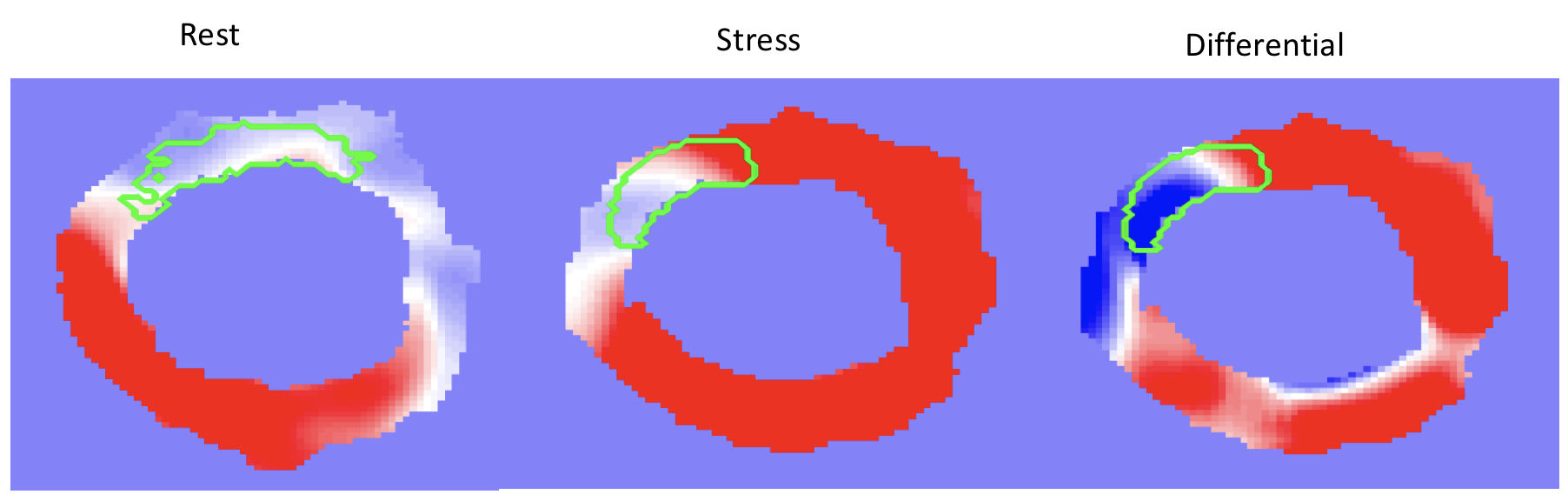}}
	\qquad \qquad \qquad 
	\subfloat[DSEC07]{\includegraphics[scale=0.3]{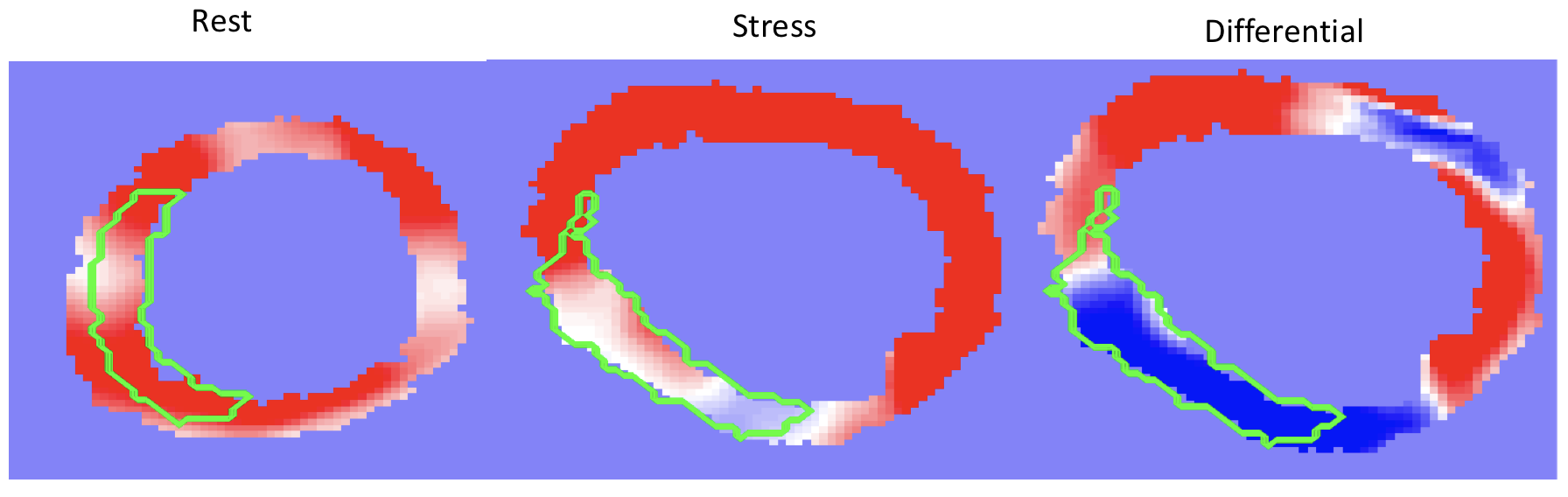}}
	\qquad \qquad \qquad 
	\subfloat[DSEC08]{\includegraphics[scale=0.3]{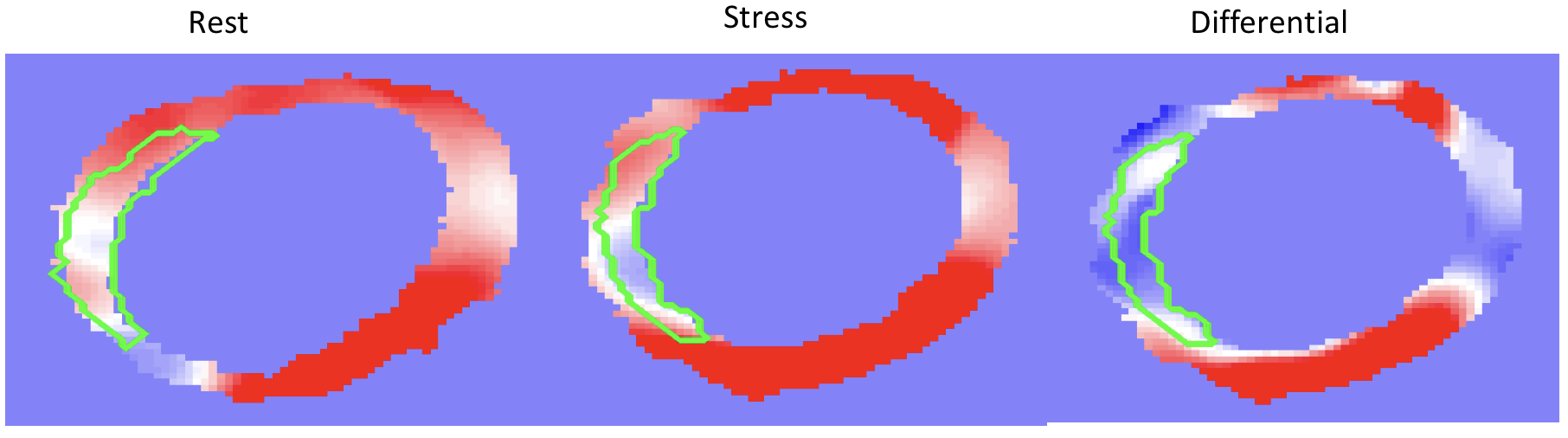}}
	\qquad \qquad \qquad 
	\subfloat[DSEC09]{\includegraphics[scale=0.3]{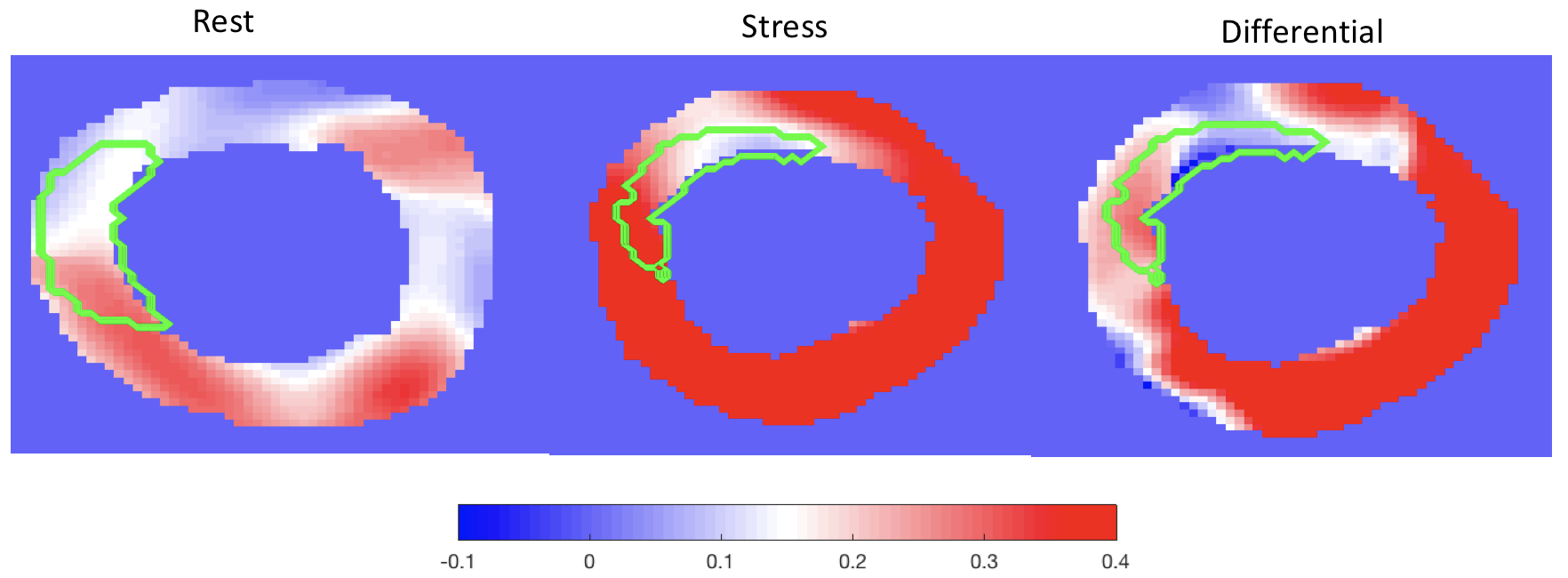}}
	\caption{\textit{Visualizations of cross-sectional Rest, Stress, and Differential strain maps for four studies}}
	\label{fig:chronic_strain_example}
\end{figure*}

Stress echocardiography is useful for detecting hidden ischemia in patients. However, currently the rest and stress images are analyzed empirically for identifying regional dysfunction, which introduces inter-observer variability. We developed a quantitative indicator that captures relative strain difference between rest and stress states that we call \textbf{differential strain}. 

We collected N=4 animal studies with chronic occlusions conducted in compliance with Institutional Animal Care and Use Committee policies. Specifically, a balloon occlusion was introduced in either the Left Anterior Descending (LAD) or the Left Circumflex (LCX) arteries. This procedure was done on day 0, and the animals was imaged on day 9. For each study, images from two animal states were acquired: rest and stressed with dobutamine. For both rest and stress image sequences, peak strains were computed using \textbf{FFD 1toF with semi-supervised learning regularization}. 

To compute differential strain, we registered ED frame of rest image sequence to that of stress image sequence, and the resulting transformation was applied to the strain map estimated from rest image sequence. Then, differential strain map was computed as the difference between registered stress strain map and transformed rest strain map. For validation, we extracted postmortem excised LV from the animal, manually traced the infarct regions for each LV slice, and reconstructed the 2D slices into a 3D surface. Using this surface, we manually traced the infarct region onto the ultrasound image. This process was illustrated in Figure \ref{fig:lad_trace}. 
\begin{table}[t]
	\centering 
	\subfloat[Rest]{
		\begin{tabular}{||l|c|c||}
			\hline
			\textbf{Studies} & \textbf{Infarct(\%)} & \textbf{Non-Infarct(\%)}  \\ [0.4ex]
			\hline\hline
			DSEC05 & 6.7 & 18.1 \\ \hline
			DSEC07 & 26.5 & 25.5 \\ \hline
			DSEC08 & 15.1 & 22.4 \\ \hline
			DSEC09 & 16.2 & 13.5  \\ \hline
			Average & 16.1 & 19.8 \\ \hline
	\end{tabular}}
	\qquad 
	\subfloat[Stress]{
		\begin{tabular}{||l|c|c||}
			\hline
			\textbf{Studies} & \textbf{Infarct(\%)} & \textbf{Non-Infarct(\%)}  \\ [0.4ex]
			\hline\hline
			DSEC05 & 10.5 & 37.5 \\ \hline
			DSEC07 & 20.0 & 32.9 \\ \hline
			DSEC08 & 16.4 & 26.6 \\ \hline
			DSEC09 & 26.4 & 43.8  \\ \hline
			Average & 18.3 & 35.2 \\ \hline
	\end{tabular}}   
	\qquad 
	\subfloat[Differential]{
		\begin{tabular}{||l|c|c||}
			\hline
			\textbf{Studies} & \textbf{Infarct(\%)} & \textbf{Non-Infarct(\%)}  \\ [0.4ex]
			\hline\hline
			DSEC05 & -0.3 & 18.1 \\ \hline
			DSEC07 & -6.7 & 10.0 \\ \hline
			DSEC08 & 3.4 & 7.5 \\ \hline
			DSEC09 & 14.8 & 34.4  \\ \hline
			Average & 2.8 & 17.5 \\ \hline
	\end{tabular} }  
	\caption{\emph{Median strain (\%) computed for the Infarct and Non-Infarct zones of the myocardium in four studies for Rest, Stress, and Differential states.}}
	\label{table:chronic}
\end{table}

Table \ref{table:chronic} presented the infarct vs. non-infarct peak strains for four studies at baseline rest (a), stress (b), and differential strains (c). In part (a) of Table \ref{table:chronic}, we observed that the change in strain between Infarct and Non-infarct zones was relatively small in the rest study: from $16.1\%$ in the infarct zone to $19.8\%$ in the non-infarct zone. However, in the stress study, we noticed that the difference in strain between infarct and non-infarct zones increased dramatically: from $18.3\%$ to $35.2\%$. This suggested that stress imaging was critical for revealing the ischemic zones in the myocardium. The computed differential strain should approximately equal to the difference between the stress image-produced strains and rest image-produced strains, but this was not always the case for the four studies. This was likely due to the mis-alignment between the rest and stress images during the registration process. Nonetheless, we were able to observe enhanced visualizations in Figure \ref{fig:chronic_strain_example}. 

\section{Conclusion}
In this work, we greatly expanded our previous work and added the ability for domain adaptation. First, we illustrated the effectiveness of our supervised neural network regularization model on synthetic data, showing improvements in both tracking and strain estimation performance. We further proposed a novel unsupervised autoencoder network with biomechanical constraints for learning a latent representation that produced more physiologically plausible displacements. We extended this framework to include a supervised loss term on synthetic data and showed the effects of biomechanical constraints on the network's ability for domain adaptation. To our knowledge, this was the first domain adaptation method for regression in the context of modern machine learning. We validated both the autoencoder and semi-supervised regularization method on in-vivo data with implanted sonomicrometers. Finally, we showed ability of our semi-supervised learning regularization approach for identifying infarcted regions using estimated regional strain maps with good agreement to manually-traced infarct regions from postmortem excised hearts. 

As for future directions, we are interested developing generative models for augmenting training of supervised displacement regularization. We believe this would greatly improve generalizability of supervised regularization learning. Furthermore, integration of learning-based regularization loss terms into the objective function of traditional tracking methods such as non-rigid registration would be interesting. We are also interested in experimenting with direct prediction of displacement field, strain map, or level of abnormality from image sequences. 


\section*{Acknowledgment}
The authors would like to thank the past and current members of Dr. Albert Sinusas' lab group who were involved in the image acquisition process. This work was supported in part by the National Institute of Health (NIH) grant numbers R01HL121226 and T32HL098069. 

\bibliographystyle{IEEEtran}
\balance
\bibliography{ref}

\end{document}